\documentclass[journal]{IEEEtran}

\IEEEoverridecommandlockouts

\usepackage{cite}
\usepackage{amsmath,amssymb,amsfonts}
\usepackage{algorithmic}
\usepackage[ruled,vlined,linesnumbered]{algorithm2e}
\usepackage{graphicx}
\usepackage{textcomp}
\usepackage{xcolor}
\usepackage{todonotes}
\usepackage{nicefrac}
\usepackage{comment}
\usepackage{wrapfig}
\usepackage{multirow}
\usepackage{makecell}
\usepackage[outercaption]{sidecap}
\usepackage{caption}
\usepackage{bm,times}
\usepackage{marginnote}
\usepackage{siunitx}
\usepackage{url}
\usepackage{tikz}
\usepackage{circuitikz}
\usepackage{tabularx}
\usepackage{svg}
\usepackage{booktabs}
\usepackage{pifont}
\usepackage{subfigure}
\usepackage{hyperref}
\usepackage{graphicx}
\usepackage{xcolor}

\newcommand{\new}[1]{\textcolor{black}{#1}}

\definecolor{new}{rgb}{0, 0, 0}

\usepackage{amsthm}

\theoremstyle{plain}

  \title{ Generalizable Motion Policies through Keypoint Parameterization and Transportation Maps}
\author{
  Giovanni Franzese,  Ravi Prakash, Cosimo Della Santina and Jens Kober
  \thanks{Giovanni Franzese,  Cosimo Della Santina and Jens Kober  are with the Cognitive Robotics, Delft University of Technology, 2628 CD Delft, The Netherlands  (e-mail:\{g.franzese; c.dellasantina; j.kober\}@tudelft.nl). Ravi Prakash is with Cyber Physical Systems, Indian Institute of Science Bangalore, India (e-mail:ravipr@iisc.ac.in).}
}

\begin{document}

\maketitle


\begin{abstract}
Learning from Interactive Demonstrations has revolutionized the way non-expert humans teach robots. It is enough to kinesthetically move the robot around to teach pick-and-place, dressing, or cleaning policies. However, the main challenge is correctly generalizing to novel situations, e.g., different surfaces to clean or different arm postures to dress. This article proposes a novel task parameterization and generalization to transport the original robot policy, i.e., position, velocity, orientation, and stiffness. Unlike the state of the art, only a set of keypoints is tracked during the demonstration and the execution, e.g., a point cloud of the surface to clean. We then propose to fit a nonlinear transformation that would deform the space and then the original policy using the paired source and target point sets. The use of function approximators like Gaussian Processes allows us to generalize, or transport, the policy from every space location while estimating the uncertainty of the resulting policy due to the limited task keypoints and the reduced number of demonstrations. We compare the algorithm's performance with state-of-the-art task parameterization alternatives and analyze the effect of different function approximators. We also validated the algorithm on robot manipulation tasks, i.e., different posture arm dressing, different location product reshelving, and different shape surface cleaning. 
A video of the experiments can be found here: \texttt{\url{https://youtu.be/bE6uOnAQBLo}}.
\end{abstract}



\section{Introduction}

One of the main appeals of robot learning from demonstrations is that it enables humans with different levels of robotic expertise to transfer their knowledge and experience about skills and tasks to the robot \cite{celemin2022interactive}. This alleviates the need to program such skills by hand, which is tedious, error-prone, and requires an expert. However, one of the long-term challenges of this approach is generalizing the learned behavior to novel situations.

By enhancing the policy with task parameterization \cite{calinon2016tutorial}, robots can generalize their learned knowledge to different variations of the same task, thus promoting scalability and data efficiency in robot learning, allowing robots to learn faster and adapt to new scenarios. For instance, a robot can be trained to clean surfaces with a reduced set of shapes, to dress an arm in a certain configuration, or to pick objects with a certain shape and place them on the right shelf. 
A versatile task parametrization to describe different situations, such as object pose, arm posture, and surface shape, is through a set of keypoints \cite{manuelli2019kpam}.  

However, the current state of the art in task parameterization struggles to scale when managing numerous task parameters, such as describing a surface as a set of points, and fails to generalize effectively for policy states that are not in close proximity to the parameterization points due to the covariate shift.

\begin{figure}[t] 
    \centering
     \includegraphics[width=0.7\linewidth]{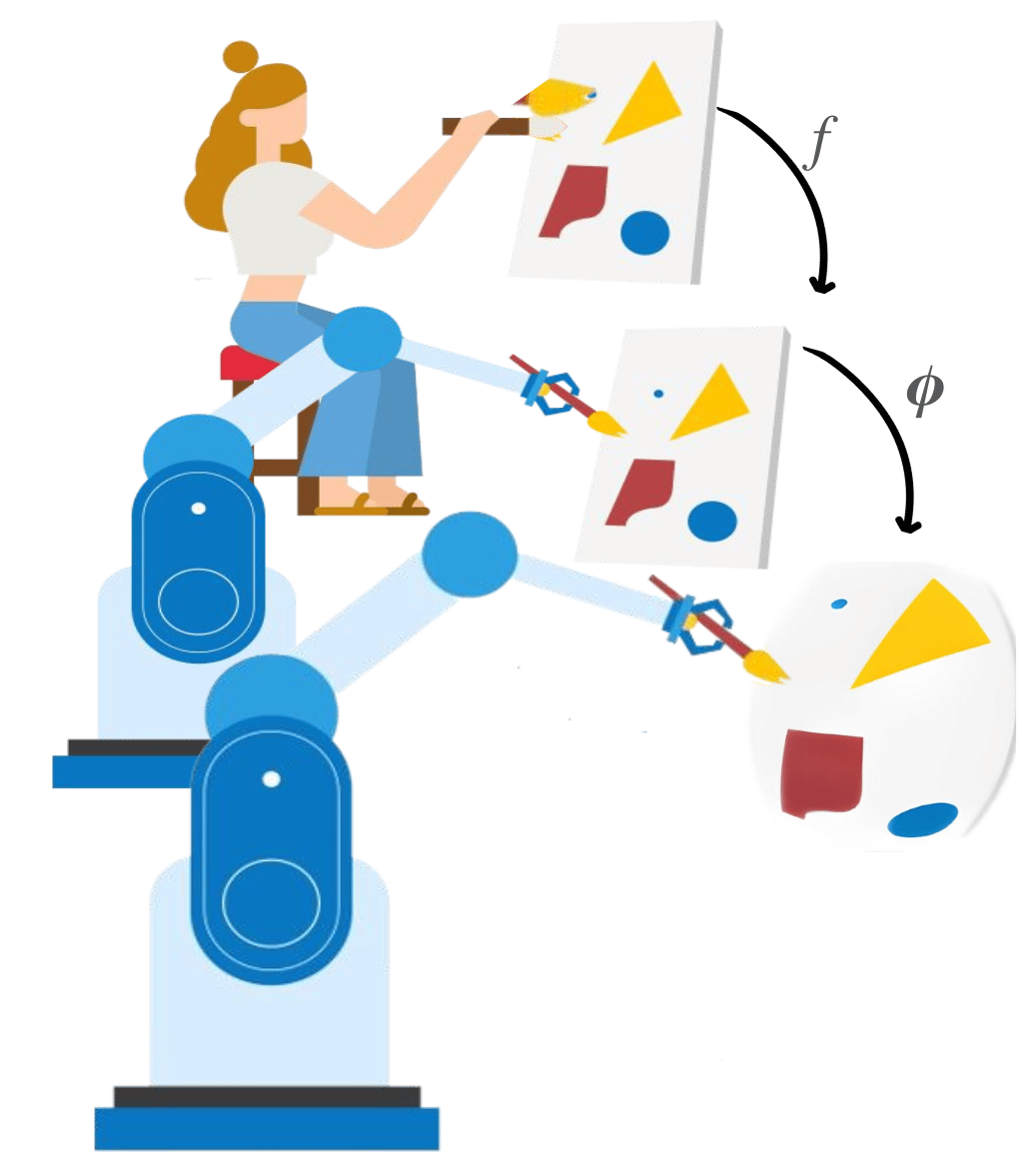}
    \caption{
    Example of Policy Transportation. The human demonstrates to the robot how to perform a task on a flat canvas. Then, the robot, when facing a new curved canvas, ``transports'' its knowledge in the new situation by adapting the end effector velocity, orientation, and stiffness to correctly adapt the drawing on the new canvas.}
    \label{fig:cover_figure}
\end{figure}

We propose the learning of a ``transportation'' map that is trained to align a set of keypoints from the original task configuration (e.g., a flat surface) to a new configuration (e.g., a curved surface) as illustrated in Fig. \ref{fig:cover_figure}. This map is then used to generalize the labels of the original policy, such as position, velocity, stiffness, and orientation.
Given that the original policy's set of labels might have been interactively adjusted or aggregated based on human feedback \cite{celemin2022interactive, franzese2021ilosa}, transforming the demonstration alone is insufficient. This limitation makes trajectory reshaping methods inappropriate for achieving the generalization goal \cite{li2023task}.

Therefore, we propose a method that not only works for the generalization of demonstrations encoded as trajectories but also if they are encoded as a set of \emph{independent} labels. The proposed algorithm will ``transport'' the original labels on the new task configuration where a new policy can be re-fitted, making the generalization algorithm agnostic to the method used for policy learning, e.g., DMP \cite{saveriano2021dynamic}, KMP \cite{huang2019kernelized}, GP \cite{franzese2021ilosa}, NN \cite{perez2023stable}, GMM \cite{khansari2011learning}, LPV \cite{figueroa2018physically}, etc.  

Moreover, our goal is to generalize out-of-distribution policies, e.g., motions that are far away from the table to clean. For this reason, it is important to define a transportation function that has desired out-of-distribution (o.o.d.) properties and is able to correctly quantify the uncertainties of the generalization. Ideally, the robot would generalize the learned skill to novel o.o.d. situations without requiring the aggregation of hundreds of thousands of demonstrations \cite{brohan2023rt}, regulate its stiffness as a function of the uncertainty \cite{franzese2023simple}, or actively ask for data in uncertain situations \cite{franzese2021ilosa}.

The next sections will highlight how this paper contributes to the field of robot policy generalization with the formalization and testing of a policy transportation theory, Secs. \ref{sec:problem_setting}-\ref{sec:convenient_choice} that can  
\begin{itemize}
    \item transport the demonstration position labels from the original task space to the new (deformed) space, see Sec.~\ref{sec:non_linear_transportation};
    \item transport the position labels, velocity labels, end-effector orientations, stiffness, and damping by exploiting the derivative of the transportation mapping, see Sec.~\ref{sec:transportation_dynamics};
    \item estimate the final uncertainty due to the reduced set of demonstrations and the estimated uncertainty in the transportation map, see Sec.~\ref{sec:transportation_uncertainty}. 
\end{itemize}

The proposed algorithm was tested on generalizing complex manipulation tasks like cleaning surfaces with different shapes,  picking and placing objects at other locations, and dressing an arm in various configurations, see Sec.~\ref{sec:experiments}.

\section{Related Works}

One classical method to generalize behavior to new situations involves task parameterization, such as the picked object location, target goal, or via points.
This idea of behavior representation and generalization in varying task configurations has been popularly achieved using Dynamic Movement Primitives (DMPs) \cite{saveriano2021dynamic} through a single or multiple demonstration per task. The DMP model consists of stable second-order linear attractor dynamics with alterable target parameters (end goal or velocities). 

An approach to adapt the DMPs via points is addressed in \cite{saveriano2019merging}, but it demands combining several DMPs for a single task. Alternatively, a roto-translation can be applied to the original dynamical systems according to the tracked frame or points in the environment and generalize the task with respect to the goal \cite{nematollahi2023robot}. 
 Approaches for modeling and generalizing demonstrations that have shown improved performances with respect to the DMPs are Probabilistic Movement Primitives (ProMPs) \cite{paraschos2013probabilistic}.
ProMPs model the distribution over the demonstrations that capture temporal correlation and correlations between the DoFs using a linear combination of weights and a set of manually designed basis functions. Adaptation to new task parameters or via points is achieved using Gaussian conditioning. While this approach allows modeling the structure and variance of the observed data in the absolute reference frame, the generalization to the new task parameters is satisfactory only within the confidence bound of the demonstration data. For example, showing many demonstrations for different goal points, the probabilistic model can be conditioned on a novel object position and retrieve the most probable trajectory that brings the robot to that final position. 
However, when learning reactive policies, i.e., a function of the state and not of the phase of the motion, the use of ProMPs is limited since the number of basis functions overgrows with the dimension of the input, limiting its applications. 

Kernelized Movement Primitives (KMP) \cite{huang2019kernelized} proposed a non-parametric skill learning formulation.
This formulation allows modulation of the recorded trajectories to new via points, obtaining the deformation of the original movement primitives given the temporal correlation of the demonstration and the via point, calculated with the kernel function. 
However, the user must specify the time and the corresponding waypoint to deform the original trajectory or rely on a heuristic that, for example, matches each waypoint with the closest point in the trajectory.

The generalization of the demonstrations, encoded as a chain of events in a graph, with respect to multiple via points, can be done using Laplacian Editing \cite{nierhoff2016spatial,li2023task}. 
It uses the Laplace-Beltrami operator, a well-known algorithm in the computer graphics community, to deform meshes \cite{sorkine2004laplacian}, to encode geometric trajectory properties and generate deformed trajectories using task constraints, i.e., new via points. The operator ensures a smooth deformation of the trajectory through the via points. 
Similarly, \cite{seker2019conditional} proposed a conditioning operation to specify the desired waypoint at a specific time of the motion. 
However, this approach is very specific for trajectory reshaping or movements that are encoded as a function of time and requires explicit knowledge of the new via point for some trajectory nodes or time instances. 

Gaussian Mixture Models (GMM) \cite{calinon2006learning,calinon2007learning,calinon2009handling} have successfully been employed in modeling demonstration, endowing with a successful generalization in its task parameterized version (TP-GMM) \cite{calinon2016tutorial}. 
Given a set of reference frames that are tracked during demonstration and execution, the central idea of TP-GMM is the local projection of the demonstrations in each of the local reference frames and encoding each model as a mixture of Gaussians \cite{calinon2012statistical,calinon2016tutorial,huang2018generalized,sena2019improving}. The local models are then fused in global coordinates, using the Product of Gaussians (PoG), and a new motion is rolled out from the resulting mixture model. 
This approach, however, requires many demonstrations to fit the model and does not scale well with the increasing number of task frames. This is because the PoG does not scale well when dealing with many reference frames and can lead to undesirable generalizations. 
Moreover, \cite{khadivar2021learning} also uses a Gaussian Mixture Regression to adopt a linear limit cycle to a nonlinear one by learning a modification of the motion amplitude as a function of the phase. However, this is very specific to phase-dependent motions, where the learned phase-dependent map would never have to be evaluated for out-of-distribution phases.

A different task-parameterized approach to generalization as formulated in \cite{perez2023learning} uses Task-Parameterized Equation Learner Networks that have as features the task parameters and the time of each demonstration and given a set of demonstrations it can regress the equations that describe the desired position as a function of time and desired task parameters, such as the desire hight of the goal. By avoiding overfitting, the method also performs well on task parameters that are outside the boundary of the observed range. However, since the parameters are considered as extra features of the regression, \emph{very far} extrapolations are not possible; moreover, the method requires multiple time-dependent demonstrations to regress the equations. We aim to relax both of these requirements and to extrapolate to extreme task parameter changes. 

In this paper, we rely on a convenient 3D keypoint parametrization since the points could be extracted from a tracked object/reference frame or a point cloud. Moreover,  we relax the need to explicitly specify the new via points for a specific node in the trajectory since the via point definition, i.e., which point of the trajectory should be moved, can be error-prone and require ad-hoc assignment algorithms between the keypoints and the trajectory.  
 \new{Yet, using keypoints is a double-edged sword. Although keypoints are a more expressive description of the task \cite{manuelli2019kpam}, if they are not moving rigidly, we cannot simply roto-translate our policy \cite{yang2024equibot, di2024dinobot}. We would have to rely on some planning \cite{huang2024rekep} or sim-2-real RL \cite{patel2025real}. 
 We are the first to contribute a nonlinear transformation in this context of imitation learning to modify the original labels and show successful zero-shot motion generalization to different surface shapes, different arm configurations in dressing, and different object locations and goal locations. }

The requirement of having a bijective map is usually used to enforce the asymptotic stability of nonlinear policy \cite{perrin2016fast} \cite{rana2020euclideanizing} or to couple the motion of two manipulators to perform bilateral teleoperation \cite{gao2021motion}. 
However, in this article, we relax the requirement on the invertibility of the map, similarly to \cite{balakrishnan2019voxelmorph}, and moreover, we use the map to transfer dynamics, orientation, and stiffness, and not only to couple the motion of two manipulators \cite{gao2021motion}. 

\textbf{Robot Reshelving:}
Authors of \cite{meszaros2022learning} propose, within the realm of robotic retail automation, to enable non-expert supermarket employees to teach a robot a reshelving task and then adeptly generalize its learned policy to accommodate diverse task situations. The generalization of the policy for varying object locations is achieved by switching between the dynamical system learned between the object and the goal frame. However, the switching strategy entails having a good prior on when to switch and all the possible implications of generating instability by suddenly changing the policy online. TP-GMM alternatives \cite{calinon2016tutorial}, solved the problem of the switching by obtaining the final GMM as the product of the relative models; however more than one demonstration is necessary to fit an informative model.

\textbf{Robot Dressing:}
 Robot dressing is a challenging task since it includes manipulation of deformable objects, and the margin of error to correctly go through the human arm is very low.
 Task parameterized dynamical systems have been applied to learn dressing tasks; for example, the dressing demonstrations w.r.t. the wrist and the shoulder of a human arm have been used to learn a dressing policy via DMP \cite{joshi2019framework}, TP-HMM \cite{pignat2017learning} and a TP-GMM \cite{zhu2022learning}. 
 
\textbf{Robot Surface Cleaning:}
Efficient and fast generalization of robotic surface cleaning has been achieved using task-parameterized learning.  
In \cite{amanhoud2020force}, the cleaning dynamics is the sum of two dynamical systems, one that learns the desired motion on the surface and another that computes the modulation term on the desired force to apply on the perpendicular direction of the surface (where the shape is known a priori). This second term is learned as a nonlinear function that allows learning larger forces in a region of the surface compared to others. 
The shape of the surface can also be estimated using the wrench measured with a force-torque sensor attached at the end-effector; for instance, 
\cite{armesto2018constraint} generalizes the polishing task on the novel curved surface by adapting the orientation and the direction of the contact force to minimize the perceived torque.

The following section formalizes the concept of policy generalization using transportation maps. 

\section{Policy Transportation}
\label{sec:policy_transportation}

\subsection{Problem Setting}
\label{sec:problem_setting}
When learning a robot policy from (interactive) demonstration, a set of desired state-action pairs is aggregated.
Throughout the paper, we denote the available \emph{policy labels} as:
\begin{itemize}
\item  The set of \( M \) Cartesian position labels \(\bm{\mathcal{X}} = \{\bm{x}_{1}, \bm{x}_{2}, \ldots, \bm{x}_{M} \}\), where \(\bm{x}_i \in \mathbb{R}^3\).
\item  The set of \( M \) Cartesian velocity labels \(\bm{\mathcal{\dot{X}}} = \{\dot{\bm{x}}_{1}, \dot{\bm{x}}_{2}, \ldots, \dot{\bm{x}}_{M} \}\), where \(\dot{\bm{x}}_i \in \mathbb{R}^3\).
\item  The set of \( M \) Cartesian orientation labels \(\bm{\mathcal{R}} = \{\bm{R}_{1}, \bm{R}_{2}, \ldots, \bm{R}_{M} \}\), where \(\bm{R}_i \in SO(3)\).
\item  The set of \( M \) Cartesian stiffness labels \(\bm{\mathcal{K}} = \{\bm{K}_{1}, \bm{K}_{2}, \ldots, \bm{K}_{M} \}\), where \(\bm{K}_i \in S^+_3\).
\item  The set of \( M \) Cartesian damping labels \(\bm{\mathcal{D}} = \{\bm{D}_{1}, \bm{D}_{2}, \ldots, \bm{D}_{M} \}\), where \(\bm{D}_i \in S^+_3\).
\end{itemize}

where \(\mathbb{R}^3\) denotes the 3-dimensional Euclidean space, \(SO(3)\) represents the special orthogonal group of 3x3 rotation matrices, and \(S^+_3\) is the manifold of \(3 \times 3\) symmetric positive semi-definite matrices. 

To develop a task parameterization that scales from pick-and-place to continuous surfaces, we track a set of environment-specific \emph{keypoints} that describe the situation. Our task parameterization comprises \(N\) tracked points in \(\mathbb{R}^3\), recorded in the demonstration scenario as the source distribution, while the corresponding moved points in the new scenario are defined as the target distribution, defined as two \textit{ordered} sets
\begin{itemize}
    \item $\bm{\mathcal{S}} = \{\bm{s}_{1}, \bm{s}_{2}, \ldots, \bm{s}_{N} \}$, where $\bm{s}_i \in \mathbb{R}^3$.
    \item $\bm{\mathcal{T}} = \{\bm{t}_{1}, \bm{t}_{2}, \ldots, \bm{t}_{N} \}$, where $\bm{t}_i \in \mathbb{R}^3$.
\end{itemize}
Here, we ask these sets to be ordered as we assume that the points of the target and source distributions are already paired. This is a reasonable assumption as this pairing can either be done manually or autonomously via state-of-the-art solutions \cite{courty2017joint,myronenko2010point}.
\vspace{1em}

\noindent\textbf{Goal}: \textit{To summarize, our goal is to find a way of learning a policy in the form of a dynamical system in the target space $\dot{\hat{x}} = g(\hat{x})$ by integrating the knowledge of the labels provided in the original space $(\bm{\mathcal{X}}, \bm{\mathcal{\dot{X}}})$ with the knowledge of the keypoints $(\bm{\mathcal{S}}, \bm{\mathcal{T}})$, giving us information on how original and target space are related.
As a secondary goal, we wish to make use of the labels $\bm{\mathcal{R}}, \bm{\mathcal{K}}, \bm{\mathcal{D}}$ in the novel task, and possibly also quantify the uncertainties of the generalization process.
}

\subsection{Proposed solution: Policy transportation}

We propose to solve this challenge via a two step process (see Fig. \ref{fig:mathematical_scheme})
\begin{itemize}
    \item First, we learn a policy transportation, mapping the labels $(\bm{\mathcal{X}}, \bm{\mathcal{\dot{X}}})$ in the target space $(\bm{\mathcal{\hat{X}}}, \bm{\mathcal{\dot{\hat{X}}}})$ in a way coherent with the keypoints;
    \item Second, we use state-of-the-art techniques to learn the dynamics $g$ from $(\bm{\mathcal{\hat{X}}}, \bm{\mathcal{\dot{\hat{X}}}})$.
\end{itemize}

The rest of the section will, therefore, focus on defining, analyzing, and learning the transportation map of step 1. We start here by introducing the general definition of the policy transportation. If the number of keypoints is larger than the size of the embedding space (3 in this work), it is, in general, not possible to match them with a linear transformation. Therefore, our policy transportation will necessarily need to be nonlinear. Let us define it as a continuously differentiable ($C^1$) map 
\begin{equation}\label{eq:policy_transporation_definition}
\bm{\phi}: \mathbb{R}^3 \to \mathbb{R}^3.
\end{equation}
Note that our secondary goal can also be achieved as a beneficial by-product of the proposed solutions. We will discuss later in this section how all the remaining labels will be transformed and the uncertainty quantified.

\subsection{Desirable properties for ${\phi}$}

\vspace{1em}
\noindent\textbf{Desired properties}: \textit{The following are desirable properties $\phi$ of\eqref{eq:policy_transporation_definition} to fulfill
\begin{itemize}
    \item[(i)] all keypoints are properly matched - i.e., $\bm{\mathcal{T}}= \bm{\phi}(\bm{\mathcal{S}})$,
    \item[(ii)] the policy transportation is a change of coordinates - i.e., $\det(J_{\phi}) \neq 0$ everywhere.
\end{itemize}
}
\vspace{1em}

The reason behind the first property should be clear, as it directly reflects the problem statement.

The reason for considering the second property is that, in general, we would like $\phi$ to transform the labels in such a way it conserves stability properties. This is not necessary to our method, but it would be beneficial since conserving stability properties may result in better performance.
Assume that $(\bm{\dot{\mathcal{X}}},\bm{\mathcal{X}})$ are samples extracted from an ideal underlying dynamical system $\dot{x} = f(x)$.
Then, we would like the policy $g$ to have the same stability properties as $f$. 
For example, if $f$ is a contractive field, then the dynamics in the new coordinates should also be contractive. \new{A necessary and sufficient condition for this is that (ii) is fulfilled \cite{tsukamoto2021contraction, kronander2015incremental, fourie2024manifold} }.

It is important to stress that of the two properties, (i) is definitely the most important, as it ensures that geometrically, the two spaces are properly matched. Instead, note that even if (ii) is not fulfilled, this will not jeopardize the stability of the learned policy. After transportation, data can be fit with a stable dynamical system, no matter the choice of $\phi$. Actually, the fact that we do not need to compute $J_{\phi}^{-1}$ is a key benefit of our method.
As such, we will impose property (i) as part of the learning process, while property (ii) will only be checked experimentally a posteriori.

\subsection{A convenient choice for the transportation map}
\label{sec:convenient_choice}

Taking inspiration from the definition of registration maps in the point cloud registration literature \cite{myronenko2010point}, we reformulate $\bm{\phi}$ as follows,
\begin{equation}
    \bm{\phi}(\bm{x}) := \bm{\gamma} (\bm{x}) + \bm{\psi}(\bm{\gamma} (\bm{x})) . 
    \label{eq:transportation}
\end{equation}
Here, \(\bm{\gamma}\) is an affine transformation and \(\bm{\psi}\) a residual nonlinear transformation. 
This way, we ensure that the nonlinear transformation \(\bm{\psi}\) only addresses the residual mismatch error remaining after the affine transformation \(\bm{\gamma}\). 
The affine transformation will, therefore, act globally, and the nonlinearities will only locally add extra deformations to fine-tune the matching process.
Fig.~\ref{fig:diffeomorphism} shows an example of such a policy transformation in action.

Before diving into the proposed two-step process for learning $\bm{\phi}$ and $\bm{\psi}$, such that they fulfill property (i), we delve into some basic analytical considerations on conditions under which property (ii) is verified.

\vspace{1em}
\noindent\textbf{Proposition}: \textit{If $\bm{\phi}$ fulfills property (i), then a necessary condition for it to also fulfill property (ii) is $\det(J_{\phi}(\bf{\mathcal{S}})) > 0$ or $\det(J_{\phi}(\bf{\mathcal{S}})) < 0$, where the inequality is to be intended element-wise. The condition becomes also sufficient if we ask it to be true across all possible choices of $\bf{\mathcal{S}}$.}
\begin{proof}
    We wish to prove that $\det(J_{\phi}) \neq 0$. The sufficient part is trivially implied by the hypothesis, as by taking all possible choices of $\bf{\mathcal{S}}$, we are sampling all the support of $J_{\phi}$.

    For the necessary part, let us assume that the hypothesis is violated, i.e. there exist two elements $s_i,s_j \in \bf{\mathcal{S}}$ such that $\det(J_{\phi}(s_i)) < 0$ and $\det(J_{\phi}(s_j)) > 0$. We define a generic continuous path $p:[0,1] \rightarrow \mathbb{R}^3$ such that $p(0) = s_i$ and $p(1) = s_j$. The function 
    \begin{equation}\label{eq:combined_path}
        \det \circ J_{\phi} \circ p:[0,1] \rightarrow \mathbb{R}
    \end{equation}
    is continuous as it is a composition of continuous functions. Indeed, $\bm{\phi}$ is of class $C^1$ by hypothesis - i.e., its Jacobian is a continuous function. The determinant of a matrix is also a continuous function, and $p$ is a continuous by construction.

    Thus, \eqref{eq:combined_path} is a continuous function starting negative in $0$ and ending positive in $1$. It follows that it must necessarily be zero at some point in $\bar{s} \in (0,1)$, thus confirming that violated hypothesis is necessary and concluding the proof.
    
\end{proof}

\subsection{Learning Transportation Maps: $\phi$}
\label{sec:non_linear_transportation}
The inference of the function is made in two steps: first, the affine transformation $\bm{\gamma} (\bm{x})$ is obtained, and then the nonlinear transformation $\bm{\psi}(\bm{\gamma} (\bm{x}))$ is fitted only on the residual error.  

\paragraph{Affine transformation}
\begin{figure*}[t!]
    \centering
    \includegraphics[width=\linewidth]{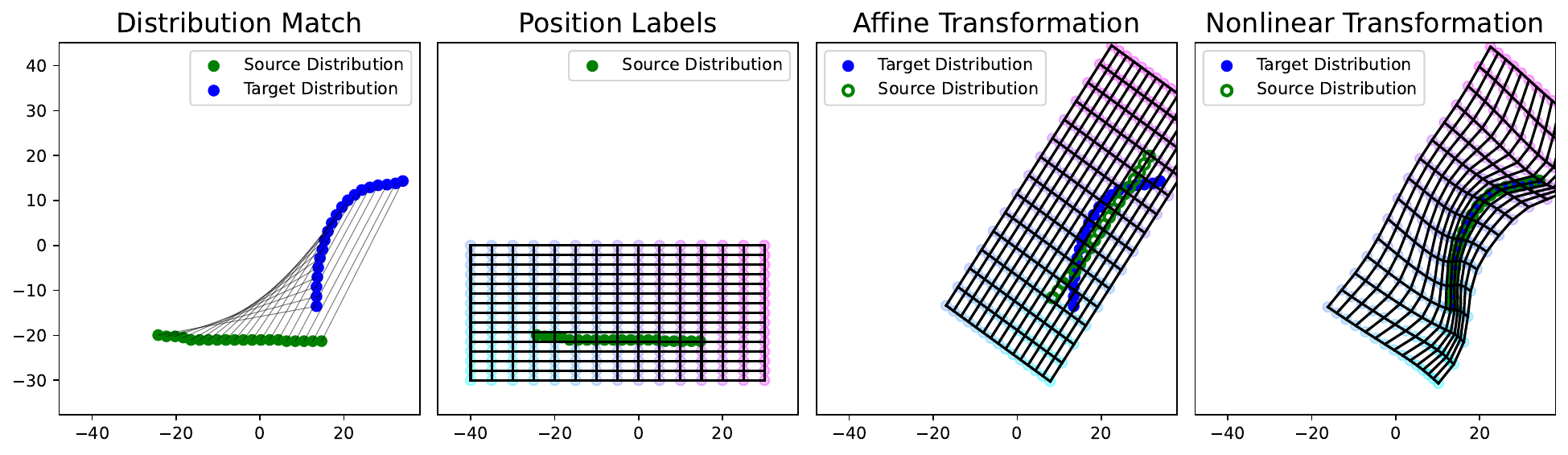}
    \caption{2D transportation.\textbf{ Distribution match} depicts the source and the target distribution correspondence used to train the transportation function. \textbf{Source Distribution} depicts a grid of points in the original space. \textbf{Linear Transformation} shows the effect on the original grid when only an affine transformation $\gamma$ is used to match source and target distribution. \textbf{Nonlinear Transportation} captures the deformation of the space when the source points are forced to match the target ones via the application of the complete policy $\phi$.  }
    \label{fig:diffeomorphism}
\end{figure*}

To fit the optimal rotation matrix between the source and the target distribution, the centered source and target distribution are used as labels for the fitting of the function $\bm{\gamma}$, i.e.

\begin{equation}
\bm{\gamma}: \mathbb{R}^3 \to \mathbb{R}^3 \text{ 
 s.t.  } \bm{\mathcal{T}}-\bm{\bar{\mathcal{T}}}= \bm{\gamma} (\bm{\mathcal{S}}-\bar{\bm{\mathcal{S}}}).
\end{equation}

where $ \bm{\bar{\mathcal{S}}} $ and $ \bm{\bar{\mathcal{T}}} $ are the centroid of the source and the target distribution, respectively. 
We can find the rotation between the two centered distributions using the  Singular Value Decomposition (SVD) imposing 
\begin{equation}
    \bm{U} \bm{\Sigma} \bm{V}^\top  = (\bm{\mathcal{S}}-\bm{\bar{\mathcal{S}}})^\top (\bm{\mathcal{T}}-\bm{\bar{\mathcal{T}}})
\end{equation}
and  the rotation matrix is defined as, 
\begin{equation}
    \bm{A}=   \bm{V} \bm{U}^ \top.
\end{equation}
However, if $ \det (\bm{A}) < 0$, the last column of $\bm{V}$ is flipped in sign, and the computation of the rotation matrix is repeated.  This ensures that the transformation is a proper rotation matrix without any reflection; see \cite{arun1987least} for more details.
Hence, the linear transformation on any point in the space can be computed as

\begin{equation}
    \bm{\gamma (x)}= \bm{A} (\bm{x}-\bm{\bar{\mathcal{S}}}) + \bm{\bar{\mathcal{T}}} . 
    \label{eq:linear_transformation}
\end{equation}
However, if $ \det (\bm{\Sigma}) = 0$, i.e., $\bm{\Sigma}$ is not full rank, it implies that the rotation is not uniquely defined and is set to the identity by default. 
Fig.~\ref{fig:diffeomorphism},c shows a linear transformation of the source and a grid of points from the original space depicted in Fig.~\ref{fig:diffeomorphism},b. 

 Having a global affine transform reduces the necessary complexity of the nonlinear function. In fact, the norm of the distance between the target and the roto-translated source is lower or equal to the norm of the distance between the target and the original source, i.e., 
$$
|| \bm{\mathcal{T}} - \bm{\gamma} (\bm{\mathcal{S}}) || \leq || \bm{\mathcal{T}} - \bm{\mathcal{S}} ||. 
$$
The proof is a direct consequence that the affine transformation transforms the original source to be as close as possible to the target, given the linear constraint.

\paragraph{Nonlinear transformation}

After fitting the linear transformation of Eq. \eqref{eq:linear_transformation}, the residual transformation is obtained by substituting the source, target points, and the fitted linear function in Eq. \eqref{eq:transportation}, obtaining that 

\begin{equation}
\bm{\psi}: \mathbb{R}^3 \to \mathbb{R}^3 \text{  s.t.  } \bm{\mathcal{T}} - \bm{\gamma} (\bm{\mathcal{S}})= \bm{\psi}(\bm{\gamma} (\bm{\mathcal{S}})).
\end{equation}

The nonlinear function $\bm{\psi}$ can be any nonlinear regressor, such as a Multi-layer Perceptron (MLP), a Gaussian Process (GP) or a Locally Weighted Translation (LWT) \cite{perrin2016fast}. However, the inductive bias given by the nature of the nonlinear function will affect the regression output when transporting points far away from the source distribution.  For example, suppose that the function is approximated with a GP with a distance-based kernel $ k $, such as a square exponential kernel (see App. \ref{sec:gaussian_process}). 
If the prior distribution is set to be a zero-mean function when making predictions in regions of the space far away from the source distribution points, the final transportation converges to just being an affine transformation, see Fig.~\ref{fig:diffeomorphism},d. Knowing the out-of-distribution (o.o.d.) properties of our policy transportation is desirable, considering that we will transport points of the policy that are not necessarily close to the point of the source/target distribution. If no keypoints are provided, the zero deformation prior is applied to the space.

\subsection{Transportation of Labels: Position and Velocity}
\label{sec:transportation_dynamics}
This paper focuses on generalizing the policy labels so that a new policy can be fitted to satisfy the new circumstances. 
Given our set of position labels $\bm{\mathcal{X}}$ from the original policy, the transported labels $\bm{\mathcal{\hat{X}}}$ are then obtained as 
\begin{equation*}
    \bm{{\mathcal{\hat{X}}}}=\bm{\phi(\mathcal{X})}.
\end{equation*}
Although the transportation map allows the transport of any point of the original demonstration in the new situation, for example, to generalize the demo on cleaning a new surface, we still have not formulated a transportation function for the velocity labels $ \bm{\mathcal{\dot{X}}}$. This is not as trivial as computing the numerical differentiation of the transported trajectories. We consider the policy labels as independent points, no longer part of a trajectory. This allows us to learn from multiple demonstrations and to change the velocity label through feedback or aggregating new data from interactive demonstrations \cite{celemin2022interactive} \cite{franzese2021ilosa}. 
Nevertheless, the partial derivative of the transportation mapping can be exploited in the velocity field generalization. 

Given the transportation function defined in the source space and projecting in the target space and by differentiating w.r.t. time on both sides and using the chain rule, we obtain the velocity labels in the transported space as 
\begin{equation}
    \bm{\mathcal{\dot{\hat{X}}}}=\frac{\partial \bm{\phi} (\bm{\mathcal{X}})}{\partial \bm{\mathcal{X}}} \bm{\mathcal{\dot{X}}} = \bm{J} (\bm{\mathcal{X}}) \bm{\mathcal{\dot{X}}}
\label{eq:velocity_propagation}
\end{equation}
where the Jacobian matrix, using the definition of Eq. \eqref{eq:transportation}, can be defined as
\begin{equation*}
   \bm{J}(\bm{x}) := \frac{\partial \bm{\gamma} (\bm{x})}{\partial \bm{x}}+ \frac{\partial \bm{\psi} (\bm{x})}{\partial \bm{\gamma} (\bm{x})} \frac{\partial \bm{\gamma} (\bm{x})}{\partial \bm{x}}
\end{equation*}
where $\frac{\partial \bm{\gamma} (\bm{x})}{\partial \bm{x}}= \bm{A}$ and $\frac{\partial \bm{\psi} (\bm{x})}{\partial \bm{\gamma} (\bm{x})}$ can be obtained using automatic differentiation of the chosen regressor. In the following sections, we will simplify notation by omitting the explicit dependence of $\bm{J}$ on $\bm{x}$.

\begin{figure*}[ht!]
  \centering
  \begin{minipage}[b]{0.45\linewidth}
    \centering
\resizebox{\linewidth}{!}{%
\begin{circuitikz}
\tikzstyle{every node}=[font=\LARGE]
\node [font=\LARGE] at (5.5,11.75) {$\mathcal{\phi}$};
\node [font=\LARGE] at (8,14.25) {${f}$};
\node [font=\LARGE] at (8,9.25) {$g$};
\node [font=\LARGE] at (10.75,11.75) {$\frac{\partial \mathcal{\phi}}{\partial x} $}; 
\draw [ -Stealth] (6.25,13.25) -- (6.25,10.5); 
\draw [ -Stealth] (10,13.25) -- (10,10.5); 
\draw[dashed] (6.75, 13.55) -- (6.75, 13.95);
\draw[dashed, -Stealth] (6.75, 13.75) -- (9.5, 13.75);

\draw (6.75, 9.8) -- (6.75, 10.2);
\draw[-Stealth] (6.75, 10) -- (9.5, 10);

\node [font=\LARGE] at (6.25,13.75) {$\bm{\mathcal{X}}$}; 
\node [font=\LARGE] at (10,13.75) {$\bm{\mathcal{\dot{X}}}$};
\node [font=\LARGE] at (6.25,10) {$\bm{\mathcal{\hat{X}}}$};
\node [font=\LARGE] at (10,10) {$\bm{\mathcal{\dot{\hat{X}}}}$};

\end{circuitikz}
}%

\caption{Mathematical Scheme of Policy Transportation.}
\label{fig:mathematical_scheme}
  \end{minipage}
  \hfill
  \begin{minipage}[b]{0.45\linewidth}
    \centering
    \includegraphics[width=0.9\linewidth]{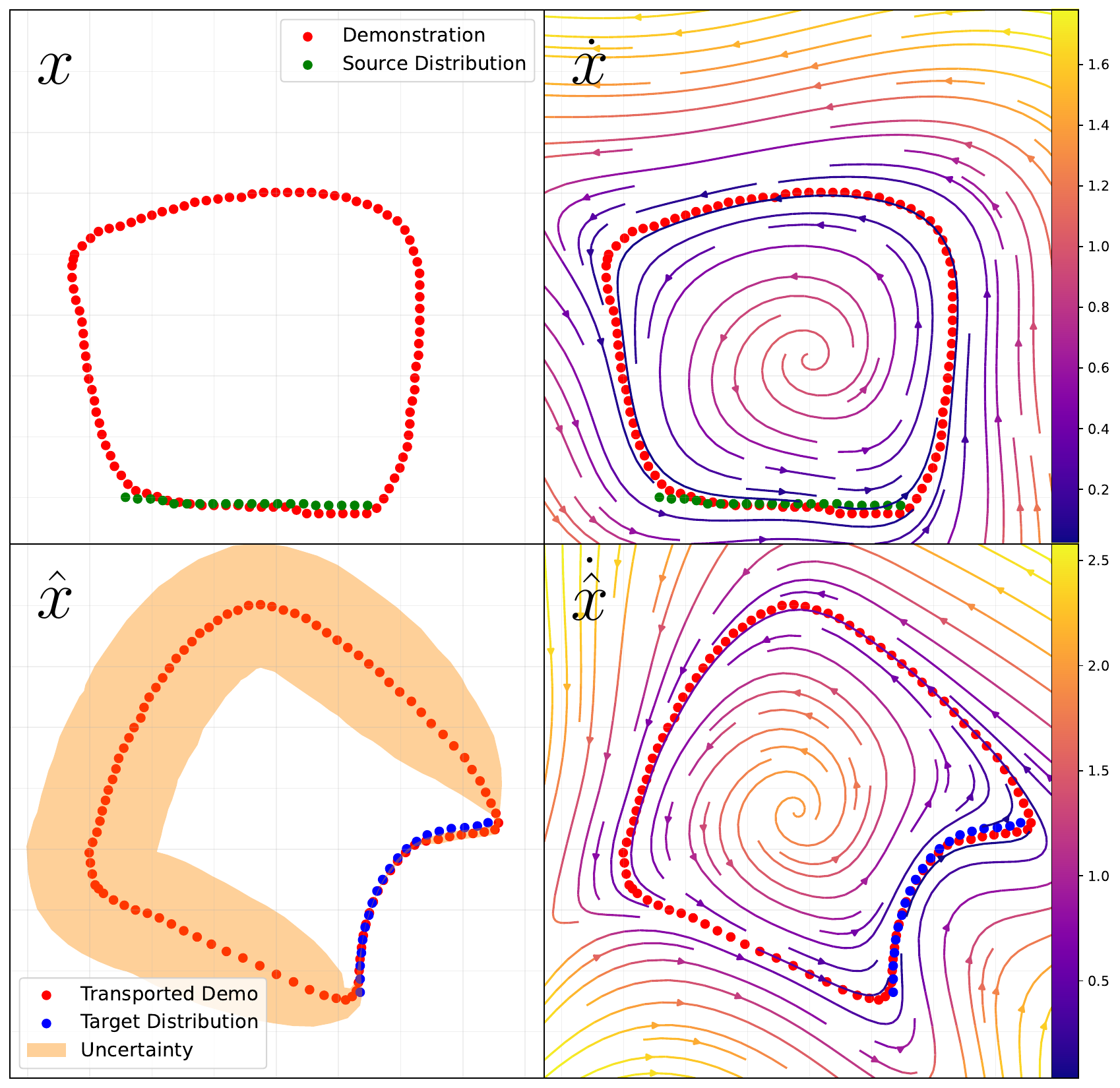}
    \caption{Graphical representation of the mathematical scheme of policy transportation. The color temperature in the fields represents the uncertainty of the prediction. }
    \label{fig:graphical_scheme}
  \end{minipage}
   \begin{minipage}[b]{\linewidth}
    \centering
    \includegraphics[width=0.9\linewidth]{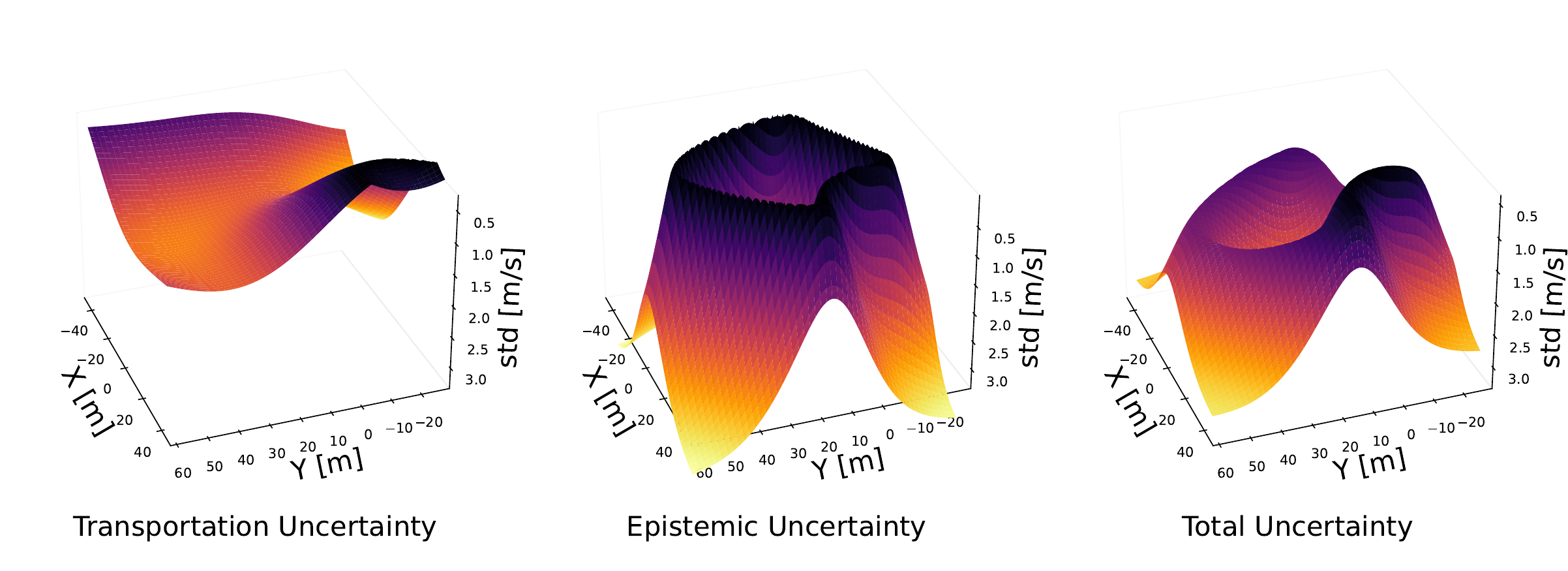}
    \caption{Standard Deviation quantification on the velocity field. \textbf{Transportation Uncertainty} 
    and quantifies the (heteroscedastic) uncertainty on the transported label (velocity) corresponding to the transported demonstration. \textbf{Epistemic Uncertainty} is the resulting model uncertainty when fitting the new policy $\hat{f}$. \textbf{Total Uncertainty} is the resulting standard deviation after computing the variance sum of transportation and epistemic uncertainties.}
    \label{fig:uncertainty_surface}
\end{minipage}
\end{figure*}

Fig.~\ref{fig:mathematical_scheme} and Fig.~\ref{fig:graphical_scheme} summarize how the transportation function $\bm{\phi}$ is used to transport the label of the original policy. We assume to have a set of labels used to fit the original dynamical system, i.e. $ \bm{\dot{x}}= f(\bm{x})$; we show how to transport these labels using the transportation function $\bm{\phi}(\bm{x})$ before fitting the new dynamical system $ g (\bm{x})$. 
If the original policy labels were sampled from a stable dynamical system, a sanity check on transportation is to ensure that the determinant of the Jacobian is larger than zero for any of the transported labels \cite{balakrishnan2019voxelmorph}; this also means that the function is locally invertible, hence locally diffeomorphic. If the map is diffeomorphic and the original labels were sampled from a stable dynamic system, they will maintain the stability property. However, 
given the new set of labels, if the task requires the motion to be asymptotically stable, then the proper regression 
\cite{figueroa2018physically}, \cite{perez2023stable}, \cite{khansari2011learning}, \cite{perrin2016fast} can be selected to enforce this requirement (even if the whole labels do not preserve this property). 
When using a function approximator that allows setting a zero-prior and evaluating out of distribution, the Jacobian converges to the average rotation matrix, i.e., $ \bm{J} \approx \bm{A}$.  Fig.~\ref{fig:graphical_scheme} shows the vector fields obtained by fitting \( f \) and \( g \) using a vanilla GP but different regressors can be chosen, like LPV \cite{figueroa2018physically} or NN \cite{perez2023stable}.

\subsection{Transportation of Labels: Orientation, Stiffness, and Damping}

When learning and controlling the Cartesian robot pose, we must also generalize the desired end-effector orientation.
Let us consider the end effector to be a vector of infinitesimal length with the base $ \bm{x_0} $ on the end effector position and pointing in the direction of the robot orientation during the demonstration, $\bm{R}$. The transportation of the tip of the vector can be obtained using the Taylor approximation of Eq. \eqref{eq:transportation}, according to
\begin{equation}
    \bm{\hat{x}_{\text{tip}}}=\bm{\phi(x_{\text{tip}})} \approx \bm{\phi(x_0)} + \frac{\partial \bm{\phi}}{\partial \bm{x}}  {\bm{\epsilon}} = \bm{\phi(x_0)} + \bm{J} \bm{R}  {\bm{\epsilon}_0}
    \label{eq:taylor}
\end{equation}
where 
$ \bm{R} \in \mathbb{R}^{3 \times 3} $ is the original robot orientation, $ {\bm{\epsilon}_0}$ is a vector with infinitesimal dimension that has zero orientation. 

Given a square matrix $ \bm{J} $, the closest rotation matrix $ \bm{J}_{\perp} \in \text{SO}(3) $ is found by projecting $\bm{J}$ onto the special orthogonal group \( \text{SO}(3) \). 
The matrix  $\bm{J_{\perp}}$ is obtained through the polar decomposition of $\bm{J}$ ensuring that $ \det(\bm{J}_{\perp}) = +1 $, thus $ J_{\perp} $ is a proper rotation matrix within $ \text{SO}(3) $.

From Eq. \eqref{eq:taylor}, it is readily apparent that the transported orientation labels of the robot end-effector become
\begin{equation}
    \bm{\mathcal{\hat{R}}} = \bm{J_{\perp}}  \bm{\mathcal{{R}}}. 
\end{equation}
 Additionally, when implementing policies on a Cartesian impedance control, the stiffness $ \bm{K} $ and the damping matrix $\bm{{D}}$ must also be transported.
The change of coordinates of the stiffness and the damping follows from  the transportation of the robot-applied force on the environment, found using Hooke's law, i.e., 
$$
\bm{\hat{F}}_s = \bm{\hat{K}} \Delta \bm{\hat{x}} = \bm{\hat{K}} \overbrace{\bm{J}\Delta \bm{x}}^{\Delta \bm{\hat{x}}} = \bm{J} \overbrace{\bm{K} \Delta \bm{x}}^{\bm{F}_s}.
$$
Hence, the generalization of the stiffness matrix becomes
$$\bm{\hat{\mathcal{K}}}= \bm{J_{\perp}} {\bm{\mathcal{K}}} \bm{J_{\perp}} ^T  $$
and following a similar reasoning for the damping matrix, we obtain, 
$$ \bm{\hat{\mathcal{D}}}= \bm{J_{\perp}} {\bm{\mathcal{D}}}  \bm{J_{\perp}} ^T ,$$
considering that the inverse of an orthogonal matrix is equal to the transpose of the matrix itself. 

\subsection{Transporting Uncertainty}
\label{sec:transportation_uncertainty}
Given the reduced set of available keypoints, some parts of the space that are not close to them will be transported with a certain degree of uncertainty. 
A probabilistic model, like a Gaussian Process (GP) or any approximator of its posterior like an Ensemble Neural Network (E-NN) \cite{deng2022deep}, will also provide the uncertainty on transported labels that are used to fit the new motion policy. 
In particular, a GP derivative is also a GP
\cite{williams2006gaussian} 
and its existence will depend on the differentiability of the kernel function, see Appendix \ref{sec:gaussian_process}. On the other hand, the uncertainty of the derivative in an ensemble neural network (E-NN) is computed as the standard deviation of the derivatives of each model.

The uncertainty quantification of the transported labels becomes essential for calculating the final uncertainty on the control variable, e.g., the velocity. In Fig.~\ref{fig:graphical_scheme}, the uncertainty is also displayed as a shaded area around the demonstration and as the ``warmness'' of the color in the vector field. 
The uncertainty of the velocity labels is due to the propagation of the original velocity labels through the derivative of the (uncertain) transportation map of Eq.~\eqref{eq:velocity_propagation}, i.e., 
\begin{equation}
    \bm{\Sigma}_{\hat{ \bm{x}}}= \bm{\Sigma}_{\frac{\partial \bm{\phi} (\bm{x})}{\partial \bm{x}}}  \bm{\mathcal{\dot{X}}}^2.
\end{equation}
given the definition of the weighted sum of Gaussian variables \cite{deisenroth2020mathematics}. 
Hence, considering that the labels are uncertain, the prediction of the resulting policy can be computed as the sum of the epistemic (due to the reduced set of policy labels) and aleatoric uncertainty (due to the reduced set of keypoints), that is
\begin{equation}
    \bm{\Sigma}_{\dot{\hat{x}}}= \bm{\Sigma}_{ \bm{\hat{f}} } + \bm{\Sigma}_{\hat{ \bm{x}}}.
\end{equation}
Fig.~\ref{fig:uncertainty_surface} depicts the \emph{transportation uncertainty} on the norm of the velocity and the epistemic uncertainty of the model $\bm{\hat{f}}$ using transported position and transported velocities labels. From Fig.~\ref{fig:graphical_scheme} and  \ref{fig:uncertainty_surface}, it is possible to appreciate that the transportation uncertainties grow when evaluating in regions that are far away from the task parameterization points since the transportation is less certain when going far away from the distribution data; on the other hand, the epistemic uncertainty grows when evaluating in points that are far from the transported demonstration. 

In conclusion, the sum of the two uncertainty fields in Fig.~\ref{fig:uncertainty_surface} grows either when we go far away from the (transported) demonstration or away from the points of the source/target distribution.

\section{2-D Simulations and comparisons}
The availability of (calibrated) uncertainties is an important feature that improves trustworthiness in deploying robot motion generalization. 
In this section, we evaluate different maps, for example, defining a Gaussian Process Transportation (GPT) for generalizing the demonstration in a 2D surface cleaning task and on a multiple reference motion generalization.
Our goal for these simulated experiments is 
\begin{itemize}
    \item to illustrate and compare how the generalization process differs when employing regressors other than a Gaussian Process or methodologies from the state-of-the-art while generalizing a cyclic demonstration that approaches and then retreats from the surface ``to clean", in Sec. \ref{sec:2Dsurface_cleaning};
    \item assess and compare the policy transportation ability to generalize in multi-reference frame tasks, measuring its performance against state-of-the-art algorithms, in Sec. \ref{sec:2Dmultiple_reference_frames}.
\end{itemize}
{ \color{new}
\subsection{2-D surface cleaning}
\label{sec:2Dsurface_cleaning}
\begin{figure*}[t!]
    \centering
    \includegraphics[width=\linewidth]{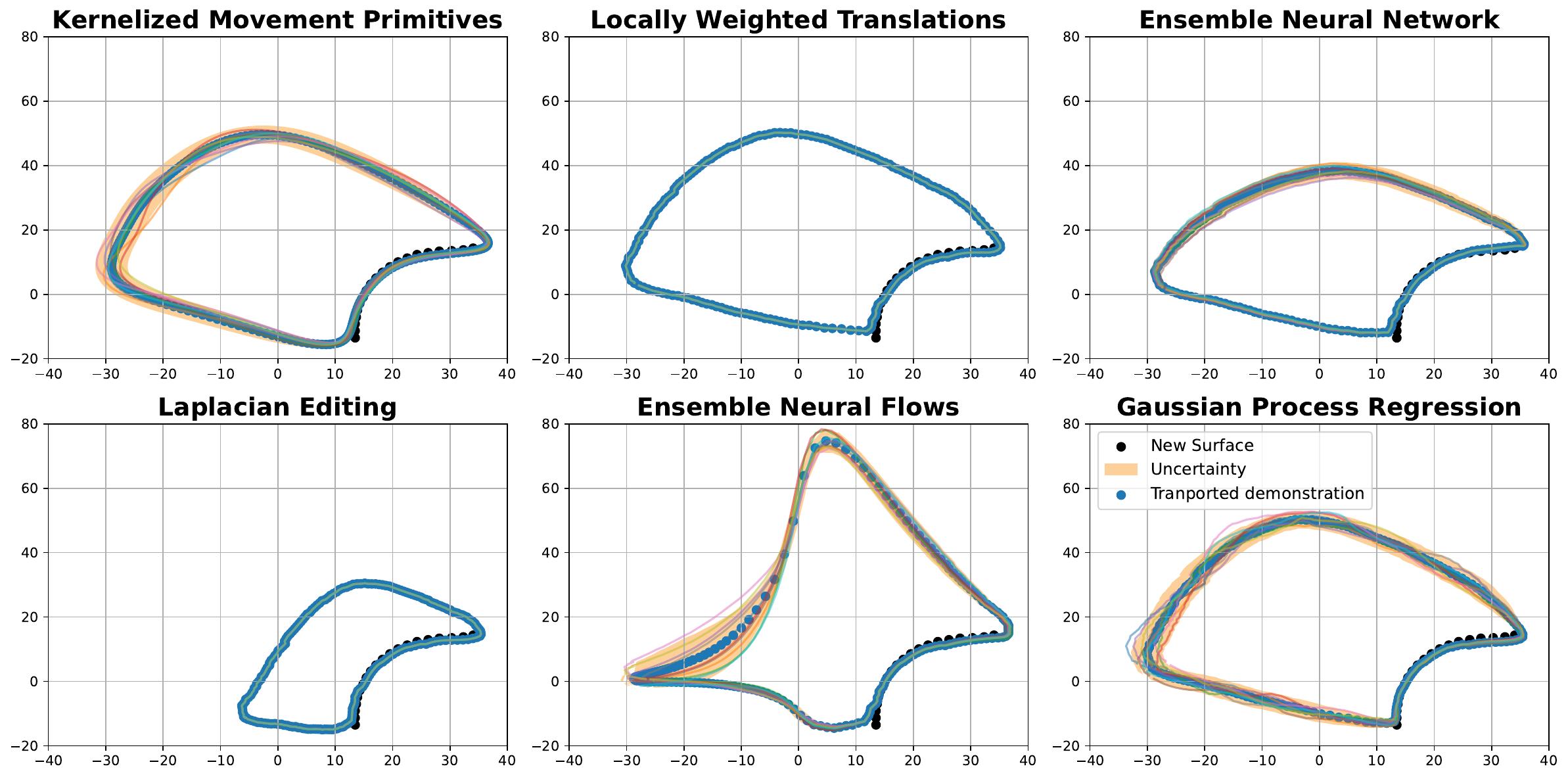}
    \caption{Qualitative comparison of the transportation of a demonstration in target space for 2D surface cleaning. 
    The colored lines are the samples of the final transported trajectory policy, i.e. $ \hat{x}= \phi(x)$, and the orange area is two standard deviations. The black curve is the 1-D surface to clean.}
    \label{fig:comparison_surface_gen}
\end{figure*}
\begin{table*}[t!]
  \centering
   \caption{Summary table of different methods used to transport trajectories to different surfaces.}
  \begin{tabular}{ccccc}
    \toprule
    \multicolumn{1}{c}{\textbf{Method}} & \multicolumn{1}{c}{\textbf{Modality}} & \multicolumn{1}{c}{\textbf{Vel. Transport}} & \multicolumn{1}{c}{\textbf{Diffeomorphic}} & \multicolumn{1}{c}{\textbf{
            Uncertainty}} \\
    \midrule
    Reshaped-KMP & linear transform + trajectory-reshaping & \ding{55}   & \ding{55} & analytical\\
    LE \cite{nierhoff2016spatial}& linear transform + trajectory-reshaping & \ding{55}  & \ding{55} & \ding{55} \\
    LTW \cite{perrin2016fast} \cite{gao2021motion} & linear transform + bijective map &  \ding{51} & \ding{51} &\ding{55} \\
    E-Flow \cite{npv2016, rana2020euclideanizing} & linear transform + bijective map &  \ding{51} & \ding{51} &estimated\\
    E-NN \cite{lakshminarayanan2017simple} & linear transform + residual map &  \ding{51} &\ding{55}  &estimated\\
    GPR & transform + residual map & \ding{51}  & \ding{55} &analytical\\
    
    \bottomrule
  \end{tabular}
  \label{tab:table_comparison_surfaces}
\end{table*}
Fig.~\ref{fig:graphical_scheme} visualizes the transportation of the given demonstration, in red, from the source to the target space, using the transportation map $\bm{\phi}$ where the non-linear component was chosen to be a GP, given the out-of-distribution prediction and the calibrated uncertainty quantification. However, other state-of-the-art function approximators can be used to fit the transportation function without loss of generality. 
To ensure a fair comparison, the mean linear transformation, i.e., $\bm{\gamma}$, is applied to all trajectories before using the different methods to perform the non-linear transportation. 
Table \ref{tab:table_comparison_surfaces} summarises the method with their properties, while Fig.~\ref{fig:comparison_surface_gen} shows the generalization of the demonstration when these methods are used. The illustrated trajectories are the result of the transportation of the original demonstration.

The figure is organized in this way:
\begin{itemize}
    \item the first column shows the result of trajectory-based reshaping methods, Kernelized Movement Primitives (KMP) and Laplacian Editing (LE);
    \item the second column shows the result by using diffeomorphic maps;
    \item the third column shows the results by using generic regression models, i.e., NN, and GPs. 
\end{itemize}
KMP \cite{huang2019kernelized}, in this study, fits the motion as a function of time, i.e. $ \mathcal{X} = f(t) $, while Laplacian editing (LE) \cite{sorkine2004laplacian, nierhoff2016spatial} considers the topology of the demonstration to be a chain, i.e., a graph where only consecutive vertices are connected with an edge or as a ring, when the demonstration is periodic, i.e. also starting and ending nodes are connected, like in Fig.~\ref{fig:comparison_surface_gen}. For both of these trajectory-based methods, every point of the source distribution is matched with the closest point of the demonstration, solving the Hungarian assignment problem. Then, each point of the trajectory, or the graph, is moved, knowing the new desired target location of the matched points by using Laplacian editing or the time-based equivalent for the transportation process, which we define as Reshaped-KMP, and it is governed by these equations: 

\[
\hat{\mathcal{X}} = \mathcal{X} + \phi(\bm{t}), \quad \text{where} \quad \phi(\bm{t_*}) = \mathcal{T} - \mathcal{S}
\]
where $\bm{t}_*$ corresponds to the time indexes in the trajectory of the selected elements to modify, and $\bm{t}$ is the complete set of time indexes of the recorded trajectory.

\begin{figure*}[h!]
    \centering
    \includegraphics[width=1\linewidth]{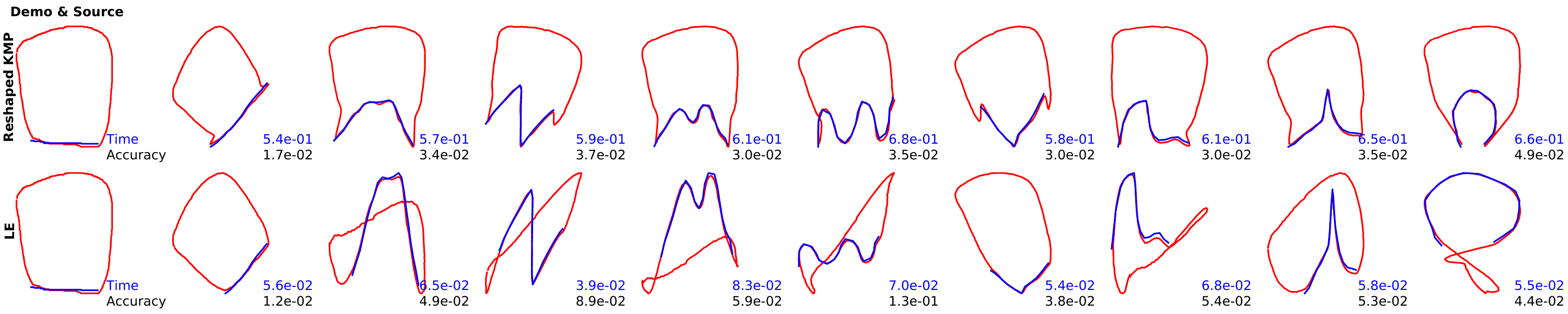}
    \includegraphics[width=1\linewidth]{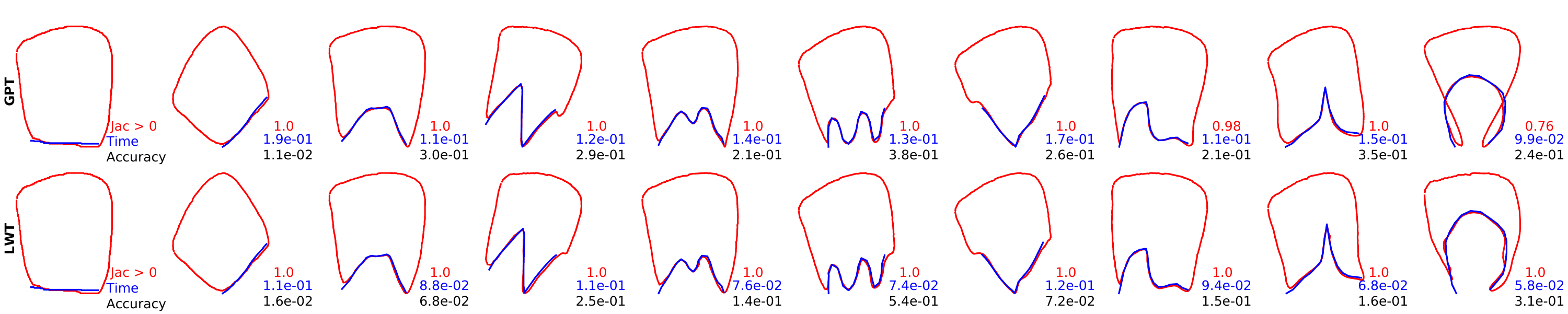}
    \caption{
  Comparisons using many more surfaces. The first two rows used trajectory reshaping methods, while the last two used the proposed transportation method with a GP and an LWT. We also report the computational time [s] to fit and transform the regressor, the accuracy in matching overlapping source points with target points, and when using transportation maps, the percentage of transported points with a Jacobian larger than zero.}
    
    \label{fig:comparisons_surfaces}
\end{figure*}

Neural Flows \cite{kobyzev2020normalizing} are bijective neural networks, i.e., flows, usually used to learn a mapping from a simple probability distribution to a more complex target distribution, and it enforces $\psi(\bm{x})$ to be a diffeomorphism; similarly a Locally Weighted Translation (LWT) \cite{perrin2016fast} also learn a diffeomorphism by consequently applying a local deformation of the space that would preserve the invertibility of the resulting total map. 
In order to estimate the uncertainties of the generalization, we train an ensemble of multiple individual models, trained independently, whose combined predictions are used to estimate a distribution on the prediction, i.e., mean and variance. 
Fig.~\ref{fig:comparison_surface_gen} depicts the mean and the uncertainty bounds of 2-$\sigma$ for the transported trajectories when using ensembles and GPs. The bounds are computed analytically for the GP from Eq. \eqref{eq:mean_gp} and \eqref{eq:variance_gp}
and the depicted GP samples of Fig.~\ref{fig:comparison_surface_gen} are drawn from the posterior distribution. 

From Fig.~\ref{fig:comparison_surface_gen}, the reader can appreciate how the GP-based methods are the only regressor with well-calibrated and unbiased epistemic uncertainty quantification and minimal mean distortion of the trajectory when transporting points far away from the source distribution. For example, the E-NN, has higher uncertainty on the right side of the trajectory, even though the points are at the same distance from the surface. The Neural Flows show an undesired distortion of the demonstration when evaluating far from the surface. Differently, the LWT performs an undistorted generalization but does not provide any measure of uncertainty.  

\paragraph*{\textbf{Quantitative Analysis}}

Figure~\ref{fig:comparisons_surfaces} illustrates how well different methods generalize a demonstrated trajectory across various surfaces. The methods compared include two trajectory reshaping techniques and two transportation methods, GPT and LWT, selected for their low out-of-distribution (o.o.d.) distortion. The figure shows that while Laplacian Editing often leads to highly distorted generalized trajectories, GPT and LWT perform similarly in terms of accuracy and computation time. For highly distorted surfaces (last column), some transported labels for the GP have a Jacobian determinant below zero, as evidenced by overlapping sections in the transported trajectory. 

}
\subsection{Multiple Reference Frames }
\label{sec:2Dmultiple_reference_frames}

\begin{figure*}[t!]
    \centering
    \includegraphics[width=\linewidth]{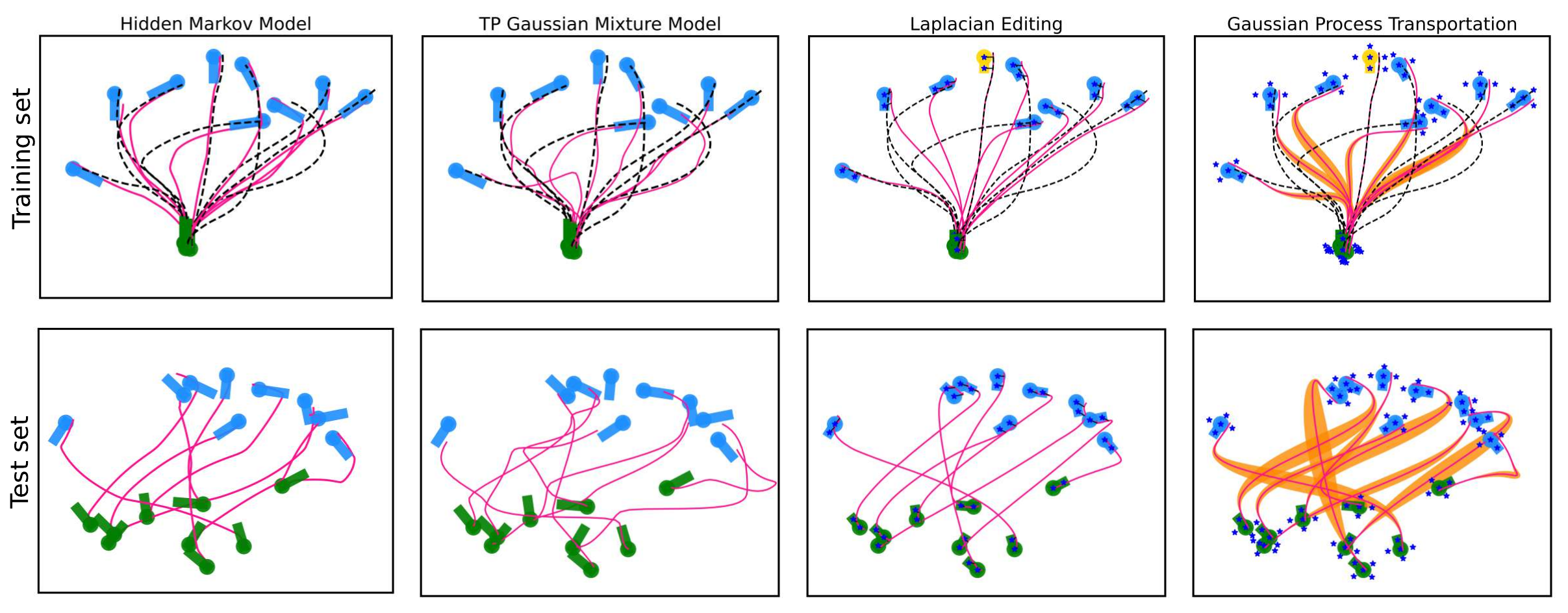}
    \caption{Qualitative comparison of multi-reference frame parameterization. Comparison between HMM, TP-GMM, LEs, and our proposed method. The first row compares the performance in reproducing the training set demonstration, depicted as a dashed black line, where both HMM and TP-GMM are trained using all nine demonstrations, while LE and GPT are only trained with one demonstration (the central one) and generalized for each of the frames. In the second row, a random perturbation is applied to each frame, and the model is queried on the most likely trajectory. For GPT, the uncertainty in the generalization is depicted with the orange areas. The blue stars in LE and GPT are the points tracked during the motion, given that our proposed method does not rely on reference frames but only on source and target points. The yellow frame belongs to the source distribution.}
    \label{fig:comparison_multi_reference_frames}
\end{figure*}

In the literature, one of the main applications of task parameterization is the generalization w.r.t. one or more reference frames. For example, if we teach a robot how to pour water into a glass, we want the robot to automatically generalize the motion w.r.t. any glass position. The task, in this particular case, can be parameterized with the location of the reference frames of each object, which is necessary to track for a successful generalization of the motion. Typically, the motion is projected in any of the reference frames, and a policy is learned w.r.t. each of the frames, leaving out the decision on the relevance of each frame for every timestep. Task-Parametrized Gaussian Mixture Model (TP-GMM) learns a GMM model for the projected demonstration for each of the frames and, during executions, the Gaussians of each frame are combined using the property that the product of Gaussian (PoG) is still a Gaussian, see \cite{calinon2016tutorial} for more details.   
Given the GMM, different control formulations are possible, for example only relying on the current state of the system, i.e., $ \bm{\Delta} \bm{x}_i= \bm{f}(\bm{x}_i)$ \cite{alizadeh2017robot} \cite{calinon2016tutorial} or by using a Hidden Markov Model (HMM) formulation that also considers the progress during the execution of the trajectory $\bm{x}_{i+1}=\bm{f}(\bm{x}_{i}, \bm{\alpha}_{i})$, where $ \bm{\alpha_{i}}$, selects the mean and variance that can be used in a tracking algorithm, such as a Linear Quadratic Regulator (LQR) \cite{calinon2016tutorial, calinon2011encoding}. However, the latent transition matrix between the different states of the HMM is unknown. They need to be estimated using a forward pass algorithm, i.e., the Viterbi algorithm \cite{calinon2011encoding}, that requires an initial guess trajectory to infer the most likely state transition that generated that initial guess and again generate the most likely motion according to the model. However, having an initial guess can be prohibitive when evaluating the movement in a novel configuration of starting and goal frames. 

Unlike these task-parameterized approaches, our proposed method does not track only the reference frame but also a set of points that are relevant to the starting and goal objects.
To describe a frame in 2D, we need at least two nonoverlapping points per frame; however, the transportation method will also scale if we extract more points; hence in Fig.~\ref{fig:comparison_multi_reference_frames}, we extracted 5 points are tracked w.r.t. each reference frame when using GPT. 
However, for the LE, we used only 2 points per frame, as the redundancy of extracted points led to ambiguity in the assignment process and decreased algorithm performance.
One of the main perks of our proposed method is the ability to generalize any motion skill generated by even only one demonstration, unlike GMM-based methods, where, to capture a meaningful mixture model, at least two diverse demonstrations need to be provided. Additionally, the GMM uncertainty of the final multi-frame model that results from the PoG does capture the uncertainty of the transportation, while, as depicted in Fig.~\ref{fig:comparison_multi_reference_frames}, the GP transportation results in growing uncertainties when transporting points of the demonstrations that are less correlated with the source-target points.
The illustrated trajectories (and uncertainties) both for GPT and LE are the result of the transportation of the single original demonstration. 
\begin{figure*}[t!]
    \centering
    \includegraphics[width=\linewidth]{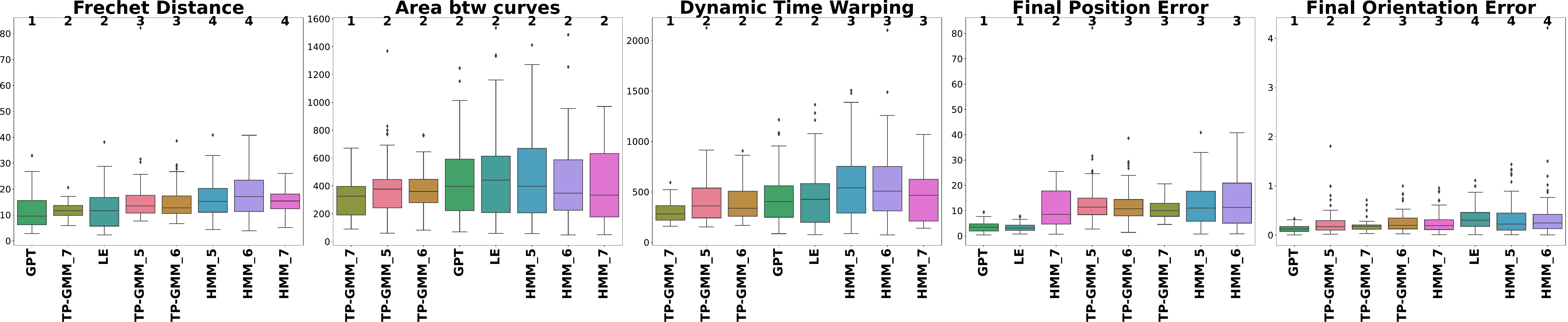}
    \caption{Box plot results and performance ranking on frame configuration from the training set. \new{The number on top of each box plot is the position of the method in the performance \textbf{ranking}.} A higher value in the ranking (where the maximum is one) implies that the method has a more statistically significant discrepancy with another method than the next method in the ranking.}
    \label{fig:box_plot_results}
\end{figure*}

\begin{figure}[t!]
    \centering
    \includegraphics[width=0.7\linewidth]{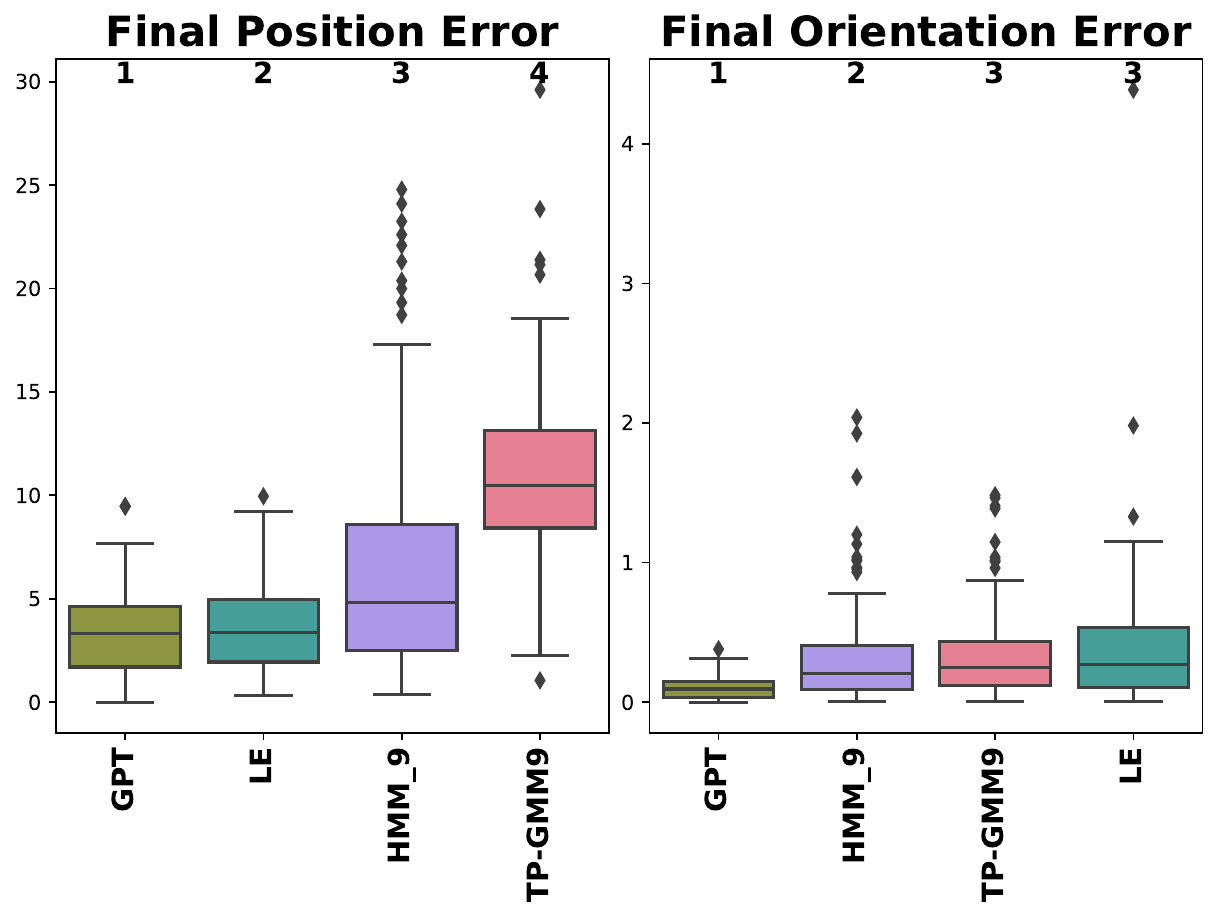}
    \caption{Box plot results and performance ranking on randomly generated frame configurations.}
    \label{fig:box_plot_ood}
\end{figure}

Fig.~\ref{fig:comparison_multi_reference_frames} highlights the discrepancy in the performance of GMM  methods on the training set and the test set. At the same time, the reproduction of a known combination of the frames results in accurate rollouts of the policies both when executing them as a dynamical system (TP-GMM)\footnote{Implementation available at \url{https://github.com/BatyaGG/Task-Parameterized-Gaussian-Mixture-Model}} \cite{alizadeh2017robot} or as an optimal tracking problem of a multi-transition Hidden Markov Model (HMM)\footnote{Implementation available at \url{https://gitlab.idiap.ch/rli/pbdlib-python/-/blob/master/notebooks/}} \cite{calinon2011encoding}. However, when evaluating the test set generated on the random reorganization of the frames, the resulting trajectories do not successfully reach the goal frame either in position or orientation. To quantify and compare the different methods, we conducted a quantitative analysis comparing the generalization on known trajectories from the demonstration set or the reaching performance on a randomly generated frame set.  
\paragraph*{\textbf{Quantitative Analysis}}
Fig.~\ref{fig:box_plot_results} shows the box plot that compares the performance of the different models, i.e., TP-GMM, HMM, LE, and GPT, on the training set. Nine demonstrations are available for different configurations of the starting and goal frames. When training the GMM models, i.e., TP-GMM and HMM, a subset of demonstration $m$ is randomly chosen from the training set and compared with the remaining $(9-m)$ demonstrations when evaluating the model in that unknown situation; the number of used demonstration is highlighted as an index, e.g., $\text{HMM\_6}$ means that we used an HMM model with six demonstrations. When training LE or GPT, only one demonstration is randomly picked from the training set and compared with the other eight unseen situations. For each model, the random selection of demonstration and comparison is repeated $20$ times. Five metrics are used to compare the rollout trajectory and the actual demonstration: 
\begin{itemize}
    \item Frechet distance that does not consider any knowledge of time but finds the maximum distance among all the possible closest pairs among the two curves \cite{jekel2019similarity};
    \item area between the two curves that constructs quadrilaterals between two curves and calculates
the area for each quadrilateral \cite{jekel2019similarity};
    \item Dynamic Time Warping (DTW) that computes the cumulative distance between all points in the trajectory \cite{jekel2019similarity};
    \item final position error, computed as the Euclidean distance between the final point of the trajectory and the rollout;
    \item final trajectory angle, that computes the approach ``docking'' angle of the trajectory. A low error in the angle distance means that the reproduced trajectory approaches the goal from the same direction as the provided demonstration in the same circumstance.   
\end{itemize}

Considering that we have many models that can behave differently according to the amount and quality of the demonstration, it is not straightforward to deduce any conclusions on which method is statistically better from the boxplot of Figs.~\ref{fig:box_plot_results} and \ref{fig:box_plot_ood}. For this reason, we run a U test, also known as the Mann-Whitney non-parametric test \cite{mann1947test}, to deduce if the distribution of results of each of the methods is statistically lower ($p < 0.05 $)  than each of the others. When computing the U test between two methods, in case of a statistical difference, the winning method gets one point. The numbers on top of the figure for each of the methods indicate the performance ranking, i.e., the method that obtained the most points when computing the U test is going to be first in the ranking. When more methods share the same position in the ranking, it simply means they were significantly better than the same number of other methods during the comparisons. 

Fig.~\ref{fig:box_plot_results} shows that for Frechet, final position and orientation error, GPT (trained with a single demonstration) performs the best. In contrast, for Area btw the curves and DTW, GPT performs equally or better than GMM and HMM models trained with five demonstrations. 

Finally, Fig.~\ref{fig:box_plot_ood} shows the box plot and ranking for the model evaluated in a test set with randomly placed frames, and GPT performs statistically better than any other method when reaching the right goal and from the right direction.

\section{Real Robot Validation}
\label{sec:experiments}
To validate the proposed method on real manipulation tasks, we selected three challenging tasks, i.e., robot reshelving (Sec.~\ref{sec:robot_reshelving}), dressing (Sec.~\ref{sec:robot_dressing}) and cleaning (Sec.~\ref{sec:robot_cleaning}),  to teach as a single demonstration and generalize it in different scenarios. These are all tasks where the training set will never be similar to the test set; for example, when dressing a human, the configuration and shape of the arm may change, and we expect the robot to generalize the behavior accordingly.

We controlled a Franka Robot using a Cartesian impedance control\footnote{\url{https://github.com/franzesegiovanni/franka_human_friendly_controllers}}.
The motion policy is learned with a non-parametric function approximation for motor learning that uses as input the position and (a belief of) time proposed in \cite{franzese2023simple}. 
The desired attractor position, orientation, stiffness, damping, and time belief update are learned as a function of the current position-time input. 
Our goal is to show how the proposed transportation theory can correctly generalize the pose, velocity, and stiffness of the robot.
The following sections will summarize the robot validation experiments.
A video of all the experiments can be found at \texttt{\url{https://youtu.be/bE6uOnAQBLo}}.

\begin{figure*}[t!]
    \centering
    \includegraphics[width=\linewidth]{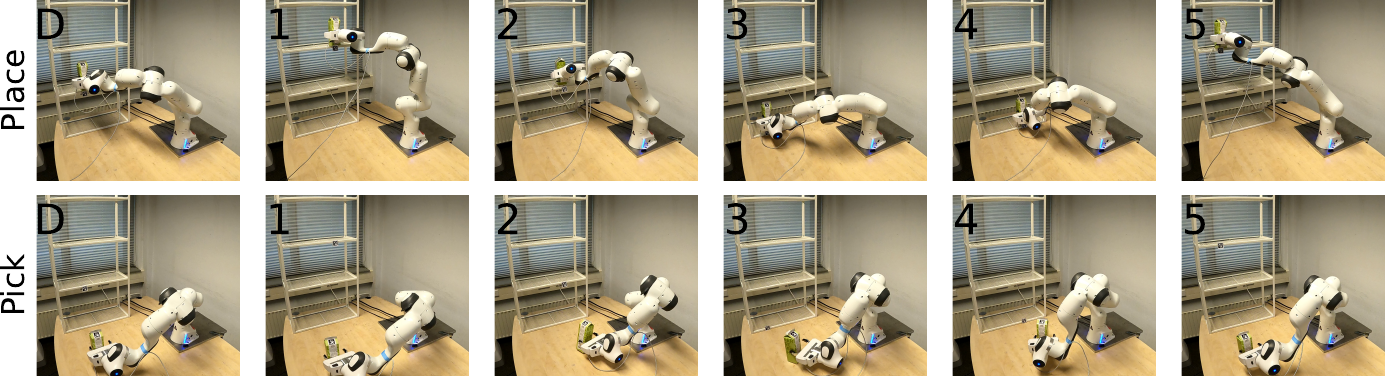}
    \caption{Generalization of the reshelving task. The first column is the robot reproduction in the demonstration scenario. }
    \label{fig:reshalving_screenshots}
\end{figure*}
\subsection{Robot Reshelving}
\label{sec:robot_reshelving}
\begin{figure}[t!]
    \centering
    \includegraphics[width=\linewidth]{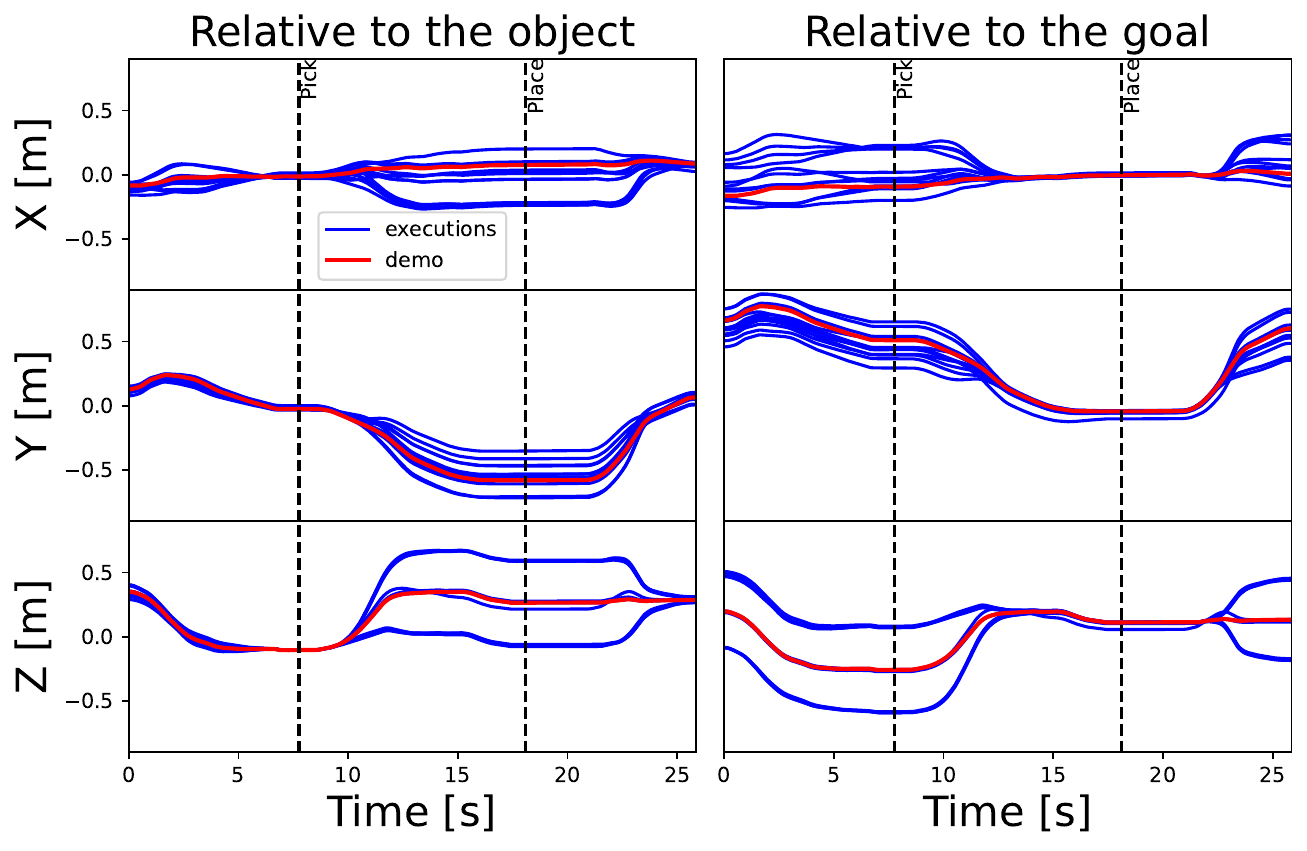}
    \caption{Relative position of the end-effector w.r.t. the initial object and the goal position during multiple generalization rollouts in robot reshelving.}
    \label{fig:reshelving_relative}
\end{figure}

Robot reshelving refers to picking an object in one location, moving it, and placing it in a desired position on a shelf. 
Our assumptions for the problem are:
\begin{itemize}
    \item one global frame movement primitive is learned from a single demonstration and transported in the different object/goal configurations;
    \item corner points of the objects and the shelf slot are tracked rather than position/orientation. To describe the 3D pose of an object, we need at least 2 non-parallel vectors; hence, we need at least 3 non-overlapping and non-collinear points. 
\end{itemize}
Fig.~\ref{fig:reshalving_screenshots} depicts the experimental setup where a milk box, with an AprilTag \cite{wang2016apriltag} on it, has to be positioned on a compartment on a shelf, also marked by another tag.
The tags are a handy way in robotics setups to track key points of objects. However, the source of the key point can be obtained with different vision models like \cite{di2024dinobot,nematollahi2022robot} without loss of generality for our proposed method. 
Before the demonstrations or execution, the robot searches for any tag in the spaces using the camera attached to its end-effector. For every tag, the transportation policy extracts a cube's center and corners with predefined side dimensions as the markers. Fig.~\ref{fig:reshalving_screenshots} shows how the demonstration for reshelving on the left of the central compartment can be generalized to any other floor, both on the left and right. We randomized the object position and orientation and the goal on the shelf ten different times, all successfully generalized. 
Table \ref{table:position_reshalving} shows the range of x,y,z, and yaw angles of the object and goal markers during the ten different executions, while Fig.~\ref{fig:reshelving_relative} depicts the relative position w.r.t.\ the object and the frame of the different rollouts; from the figure it is possible to appreciate how the execution lines converge on the (initial) object position when picking and on the goal position when placing the object. 

\begin{table}[t]
\caption{Range of Variability for Object and Goal Frames.}
\centering
\begin{tabular}{|c|c|c|c|c|}
\hline
\textbf{Frame} & $x$ [m] & $y$ [m]& $z$ [m]& yaw [deg] \\
\hline
\textbf{Object} & 0.225 & 0.366 & -  & 94.6\\
\hline
\textbf{Goal} & 0.337 & 0.036 & 0.675 & -\\
\hline
\end{tabular}

\label{table:position_reshalving}
\end{table}

\begin{figure*}[t!]
    \centering
    \includegraphics[width=\linewidth]{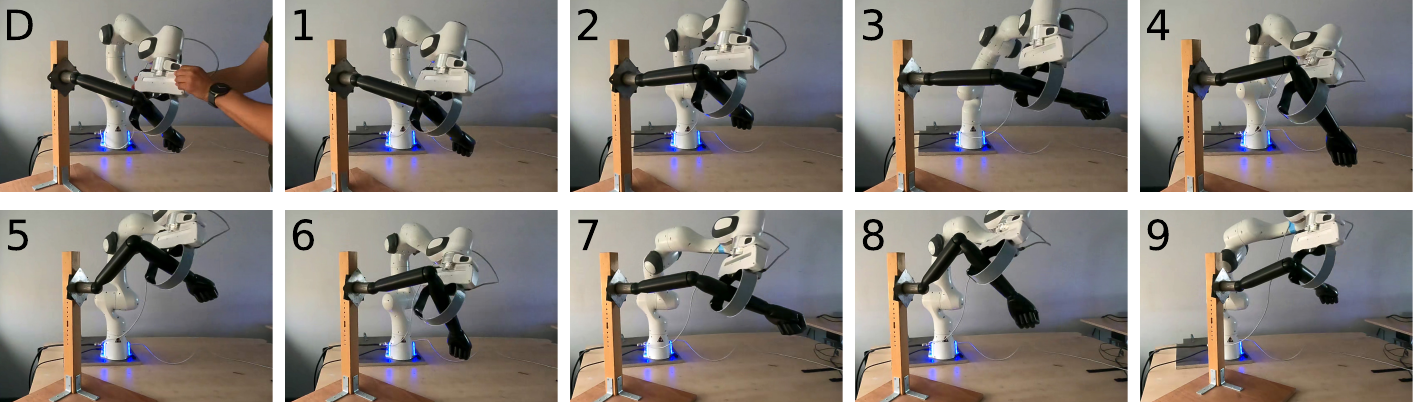} \\
    \vspace{2mm}
    \includegraphics[width=\linewidth]{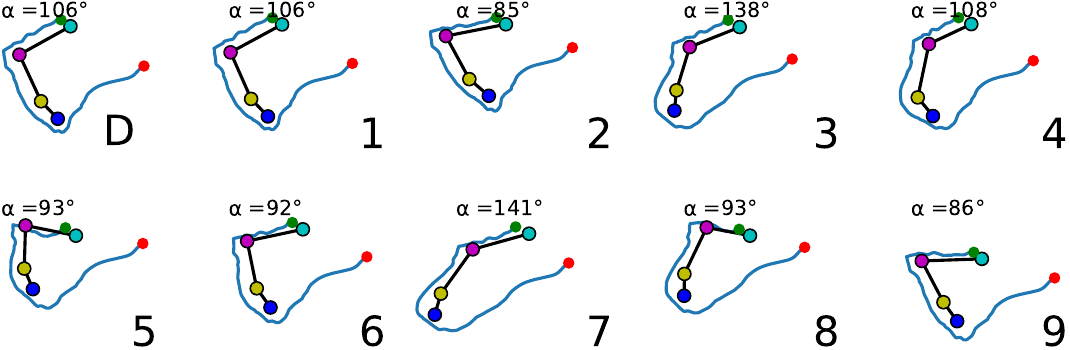}
    \caption{Dressing policy generalization. Cyan marker is the shoulder, magenta is the elbow, yellow is the wrist, and blue is the hand. The blue rollout end-effector trajectory starts from the red dot and finishes with the green dot. $\alpha$ is the angle of the elbow; the smaller the angle, the more complicated the generalization would be. }
    \label{fig:dressing_tasks_snapshots}
\end{figure*}

\subsection{Robot Dressing}
\label{sec:robot_dressing}
 The task of dressing is a primary task in elderly care. It consists of pulling a deformable sleeve over the posture of a human arm. 
Complicated motions need to be executed by the robot to increase the dressing success rate, i.e., reaching the shoulder without getting stuck or exercising too large force on the arm. Fig.~\ref{fig:dressing_tasks_snapshots} depicts the robot experimental setup where an articulated mannequin is posed in different shoulder positions and arm configurations. Four AprilTags \cite{wang2016apriltag} are glued on the arm, shoulder, elbow, wrist, and hand, captured by the camera on the robot wrist at the beginning of each demonstration/execution. From the markers, only the center point is extracted. The piece of cloth is pinched in the end effector by the user before starting the experiments. We leverage the assumption that the pose will not change during the demonstration; however, it is worth mentioning that the arm structure is not fixed on the table, so if the generalization is not good and the robot maliciously touches the arm, the resulting displacement would result in unsuccessful dressing. Only one demonstration was given to the robot. Then, the arm was reset for a different range of x,y positions of the shoulder and configuration of the arm. The ranges of variation of the task parameters are $\Delta x_{\text{shoulder}} = 0.122 \text{ [m]} $, $\Delta y_{\text{shoulder}} = 0.259 \text{ [m]} $, $\Delta \alpha = 56 \text{ [deg]} $, where $\alpha$ is the angle that the elbow intercepts with the connecting line between the shoulder and the wrist. A fully stretched arm (easy pose to dress) has $\alpha = 180$, and when the hand touches the shoulder (impossible pose to dress) $\alpha = 0$. The policy transportation was able to generalize the policy for every requested arm configuration. 
 
\subsection{Robot Surface Cleaning}
\begin{figure*}[t!]
    \centering
    \includegraphics[width=\linewidth]{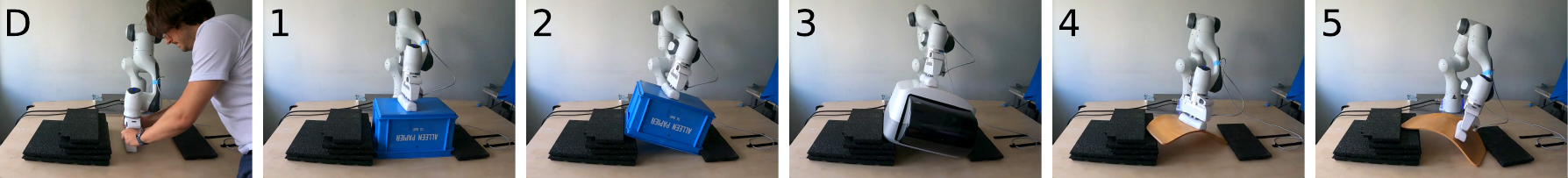} \\
    \vspace{2mm}
    \includegraphics[width=\linewidth]{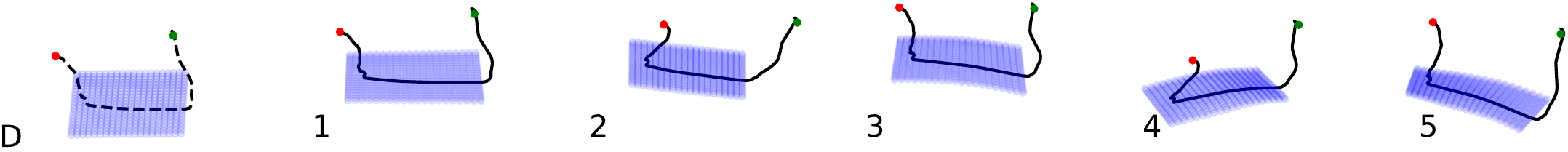}
    \caption{Point cloud, demonstration and rollouts of the generalized motion in cleaning tasks.}
    \label{fig:cleaning_pointclouds}
\end{figure*}

\begin{figure}[h!]
    \centering
    \includegraphics[width=1\linewidth]{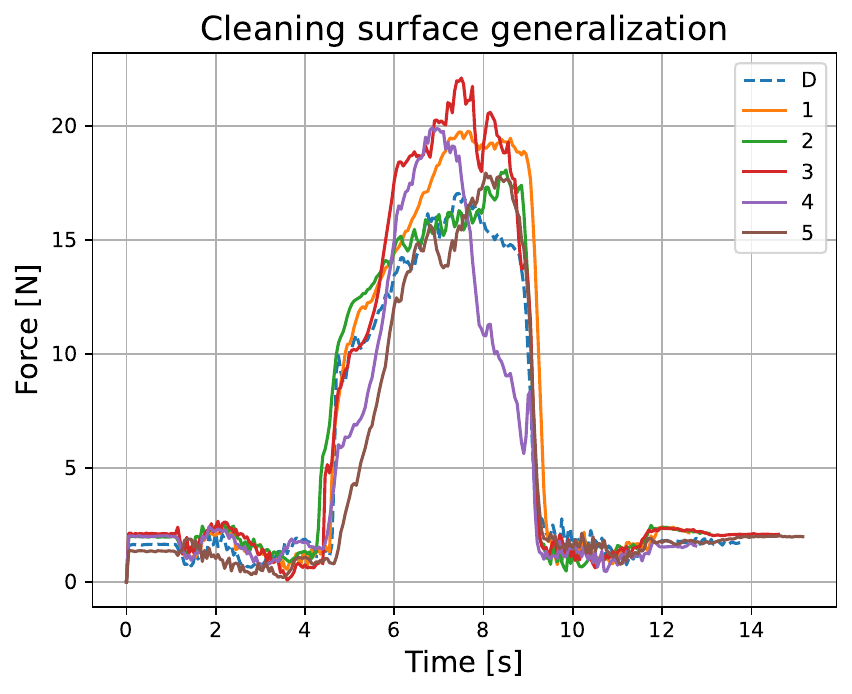}
    \caption{Norm of the force perceived [N] from the end-effector when executing the transported dynamics on the new surfaces. }
    \label{fig:force_norm}
\end{figure}
\label{sec:robot_cleaning}

Surface cleaning/grinding tasks require robots to not only track surface shapes but also apply the right amount of force for successful cleaning/grinding. Robotic cleaning or grinding involves automated machines equipped with specialized tools to perform cleaning tasks. 

In this experiment, we want to show that 
\begin{itemize}
    \item we can learn a general policy that may involve polishing phases and free movement phases;
    \item we do not need any force sensors to align to the surface;
    \item the surface is unknown, and only an ordered point cloud is obtained from the camera sensors. 
\end{itemize}
\new{The point cloud is obtained from a RealSense D435 camera. The user selects the relevant area, after which the points are smoothed along the z-axis and organized into a grid on x and y coordinates of dimension $20\times20$ where the corners are specified by a user input. This ensures that the source and target distributions have the same number of points, allowing for accurate matching and minimal estimation noise.}    
One of the main advantages of the proposed method is that it does not need to reconstruct the surface but only learns the map from the source to the target point cloud.
The transportation between the (smoothened) source and the (smoothened) target surface point cloud is modeled using a Stochastic Variational Gaussian Process Transportation (SV-GPT) to generalize the demonstrated policy position, orientation, and stiffness profile for a successful cleaning task. Given the large number of points in the source and target point cloud, i.e., 400 points, using a reduced set of inducing points, i.e., 100, makes fitting the transportation model more computationally efficient, see Appendix~ \ref{sec:gaussian_process}.

Fig.~\ref{fig:cleaning_pointclouds} depicts the teaching of a cleaning task on a flat surface and the generalization on different heights, tilted, and curved surfaces that belong to common objects. The lower row shows what the robot perceives of the environment; the blue dots in space are the source distribution, recorded before giving the demonstration (depicted as dashed line), and target distributions recorded before executing the rollout transported policy (depicted a solid line).
Fig.~\ref{fig:cleaning_pointclouds} also highlights how the rollouts follow the shape of the surface, showing a successful generalization of the robot position and orientation. 
As previously stated, no external force-torque sensor is used to adapt the orientation of the end-effector to the tangential direction of the surface. However, an observer of the applied external force between the robot and the surface is estimated from measured torques in the joints. Fig.~\ref{fig:force_norm} depicts the estimated norm of the force exchanged with the surface, where the same increasing/decreasing trend is captured on the different surfaces. 

\begin{figure}[ht!]
    \centering
\includegraphics[width=\linewidth]{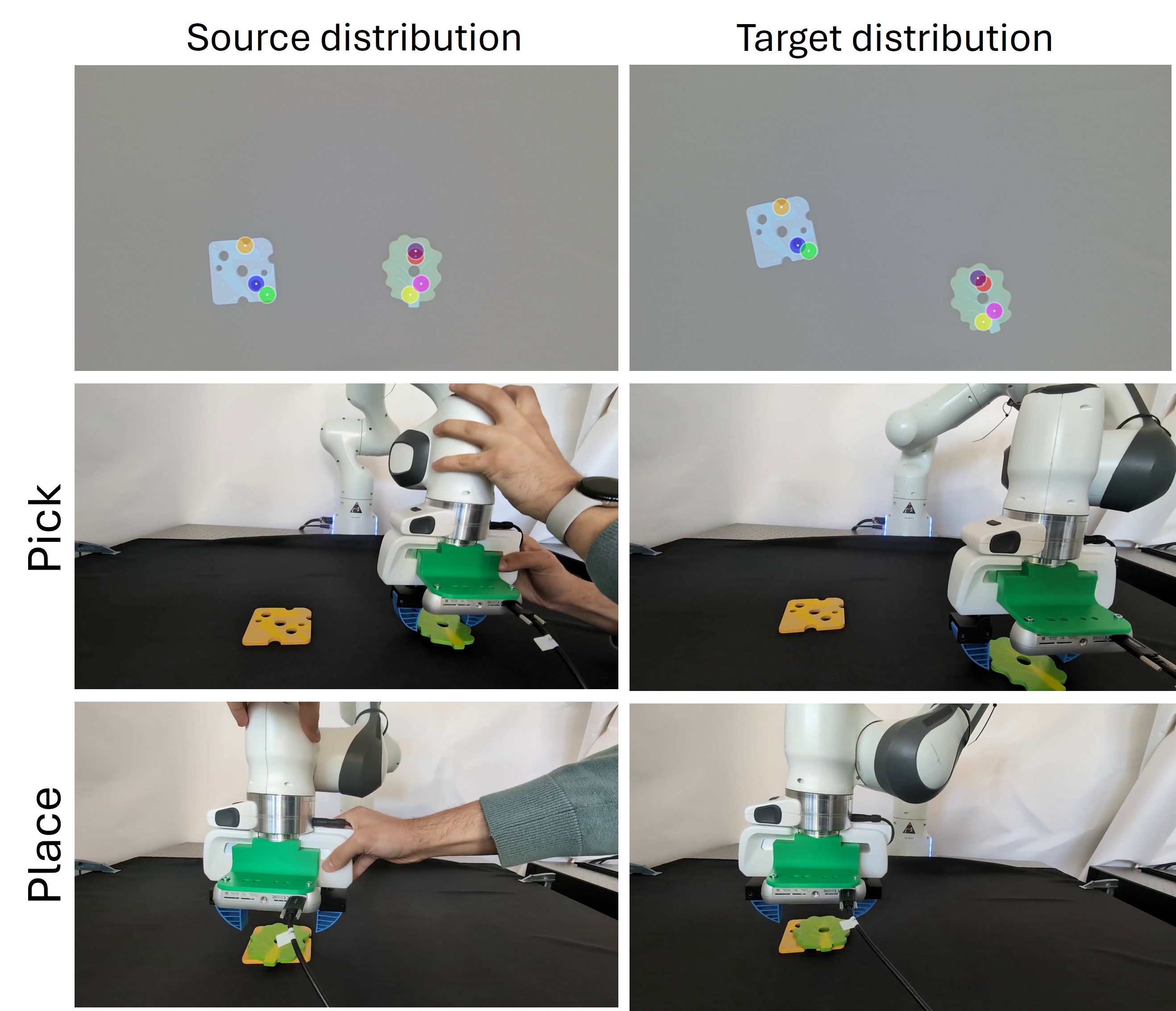}
    \caption{\new{Food Stacking Task. The single demo is displayed on the left, and the motion generalization is on the right. The source and the target keypoints are automatically detected using DINO features correspondences \cite{dino_features}. The 3D location of the keypoint correspondence is obtained using the stereoscopic depth map. }}
    \label{fig:burger_assembly}
\end{figure}
\new{
\subsection{Towards policy transportation using DINO features}
Relying on key points extracted from fiducial markers is not always practical in real-world applications. Nevertheless, given the raw camera input and an estimated depth, we can leverage new foundation models to automatically extract and find the corresponding features \cite{dino_features}. This experiment is similar to what was shown in DINObot \cite{di2024dinobot}. However, we are not only roto-translating the policy, but we also apply the non-linear transformation proposed in this paper. 
Fig.~\ref{fig:burger_assembly} shows the source, recorded before giving the demonstration, and the target image on a new out-of-distribution configuration of the objects. A stereoscopic depth map is used to estimate the 3D location of each feature, so as to generate the necessary keypoints to perform the generalization. We ask the model to find a maximum of 7 key points, and a threshold of 0.2 was applied to the saliency maps, retaining only features with higher attention values; this avoids finding features on the table. 
We moved the objects around the workspace, always with the (toy) cheese on the left and the (toy) lettuce on the right, and used a transportation policy with a GP to perform the generalization of the original position and orientation labels. Despite very successful results in generalizing in 10 different configurations, we identify poor orientation generalization when swapping the location of the objects, i.e., lettuce on the left and cheese on the right side of the table. On the computational side, the main bottleneck is the DINO feature extraction, which, on the current implementation \cite{dino_features} on an RTX4060, can take almost a minute to find the keypoint correspondence. Moreover, the depth estimation turned out to be the brittle component of the pipeline when working with objects that are not flat. 
}

\section{Limitations and Open Challenges}
Despite the successful application of the proposed policy transportation on different challenging tasks, we can foresee some limitations and future challenges to improve the applicability and have a broader impact. 

For example, we assume knowing the matching between the points of the source and the target distribution. However, in many complex scenarios, this limitation can be problematic, and some different pre-processing algorithms, such as optimal transport\cite{luo2023optimal} or iterative closest point (ICP), or Coherent Point Drift \cite{myronenko2010point} need to be adopted to perform the matching. Additionally, semantic matching can increase cross-domain generalization, for example, by adapting the reshelving strategy to a completely different shelf type or adjusting the dressing policy from an adult to a baby arm or to manipulate different food shapes or reshelving boxes of different dimensions. 

\new{Moreover, in the current implementation and experimentation, we assume full observability of the keypoints and can estimate their location with high accuracy. We foresee exciting directions in dealing with partial observability and different sensor noise, in particular when working with objects with very irregular shapes and estimating the keypoint location using the depth cameras.} 

\new{Another assumption of the developed method is that it deals with static environments, i.e., the target distribution does not change during the policy's rollout. However, this assumption can fail when dealing with the reshelving of moving objects or when trying to dress real humans that will probably move before and during the interaction \cite{li2021provably, zhu2024you}. Nevertheless, supposing to know the state of the target distribution, the transportation policy can be updated inexpensively online since only the label $\hat{y}$ of Eq. \eqref{eq:transportation} and the (cheap) linear transformation needs to be recomputed. However, the fitting of the transported policy $\hat{f}$  makes it challenging to perform the generalization online.} 

Finally, given that in complex scenarios, the generalization may be inaccurate, the use of interactive human corrections may increase the resulting manipulation performance \cite{celemin2022interactive}.  However, changing the generalized policy opens the question of whether interactive corrections should be propagated back to the original policy labels and how. Additionally, in case many source distributions/policies are recorded, selecting the best one, according to some similarity to the current observed target distribution, can open exciting developments of scaling the proposed theory to work on a big database of keypoint parameterized demonstrations. 

\section{Conclusions}
In this paper, we address the prominent but challenging problem of policy generalization to novel unseen task scenarios. We formulate a novel policy transportation theory that, given a set of matched source and target points in the task space of the robot, regresses the function that, most likely, would match the source and target distribution. Additionally, we showed how the same transportation function and its derivatives can be exploited to transport the original policy dynamics, rotation, and stiffness while retaining uncertainties in the process. 

The same transportation algorithm was tested with different state-of-the-art regressors and compared with different generalization methods, showing how, even with only one demonstration, it results in better or comparable motion generalization. 

We validated the proposed approach on a Franka Robot, testing it on four different tasks: product reshelving, arm dressing, and surface cleaning, and table-top pick-and-place. These various tasks were never tackled together by the same generalization algorithm, and they were usually performed with ad hoc solutions, for example, to keep a constant force when cleaning a surface. Despite this, our policy transportation algorithm performed successfully in all of them. The tracking requirements were satisfied using fiducial markers or directly the point cloud estimated with a stereo camera. 
\new{Future development will have to focus on scaling the manipulation generalization using large, unmatched, and partially observable point clouds of complex, deformable, and moving objects while allowing the use of human feedback in the fine-tuning of the resulting policy, asking for human attention when facing high-uncertainty generalization regions, or actively searching for missing keypoints that would reduce the total transportation uncertainty.}

\bibliographystyle{IEEEtran}
\bibliography{reference} 
\appendices
\renewcommand{\thesection}{\Alph{section}}  
\section{Gaussian Process Regression}
\label{sec:gaussian_process}
A Gaussian Process is a collection of random variables, any finite number of which have a joint Gaussian distribution. To fit a Gaussian process, we start with a prior distribution over functions. The prior is typically specified as a mean function and a kernel function. The prior distribution represents our beliefs about the functions before observing any data. For example, when learning a dynamical system, it is safer to have a zero mean prior, such that the robot does not attempt to do any movement if there is no significant evidence from the human demonstration. 

The posterior distribution is used to make predictions or perform inferences, and assuming Gaussian likelihood, the mean and variance predictions are
\begin{equation}
    \bm{\mu}=\bm{K}_{\bm{X}_*,\bm{X}}(\bm{K}_{\bm{X},\bm{X}}+ \sigma^2_n \bm{I})^{-1} \bm{y}
    \label{eq:mean_gp}
\end{equation}
\begin{equation}
    \bm{\Sigma}=\bm{K}_{\bm{X}_*,\bm{X}_*}-\bm{K}_{\bm{X}_*,\bm{X}}(\bm{K}_{\bm{X},\bm{X}}+ \sigma^2_n \bm{I})^{-1} \bm{K}_{\bm{X},\bm{X}_*}
    \label{eq:variance_gp}
\end{equation}

where $\bm{X}_* \in \mathbb{R}^{n \times m}$ are the evaluation inputs and $\bm{X} \in \mathbb{R}^{N \times m} $ and $\bm{y} \in \mathbb{R}^{N \times l}$ are the training inputs and outputs; the correlations $\bm{K}_{\bm{X},\bm{X}} \in \mathbb{R}^{N \times N} $, $\bm{K}_{\bm{X}_*,\bm{X}} \in \mathbb{R}^{n \times N} $, $\bm{K}_{\bm{X}_*,\bm{X}_*}\in \mathbb{R}^{n \times n} $ are the correlation matrix between every input training point, between testing and training points and between testing points. Every entry of this matrices is computed using a kernel function, for example, the squared exponential  kerneldefined as

\[
k_{\text{SE}}(\bm{x_i}, \bm{x_j}) = \sigma_p^2 \exp \left( -\frac{(\bm{x_i} - \bm{x_j})^2}{2 \ell^2}\right),
\]
where $\bm{x_i}, \bm{x_j}$ are any pair of training-training, testing-training, or testing-testing data points; 
$\sigma_p$ and $\ell$ are the kernel hyperparameters and, together with the likelihood noise $ \sigma_n$ are optimized with a maximization of the log-likelihood of the given data \cite{williams2006gaussian}. \new{In the context of this paper, when dealing with multi-output GPs, we assume the same smoothness, prior noise and likelihood noise among the models, hence we compute $(\bm{K}_{\bm{X},\bm{X}}+ \sigma^2_n \bm{I})^{-1}$ once for all the three models and during inference the vector  $\bm{K}_{\bm{X}_*,\bm{X}}$ is also the same for the three outputs. }

Moreover, given the computationally expensive inversion of the covariance matrix that grows with the number of data points, optimizing kernel hyperparameters and inference becomes prohibitive when having big datasets. Therefore, variational inference aims to approximate the true posterior \( p(\bm{f} | \bm{X}, \bm{y}) \) with a simpler distribution \( q(\bm{f}) \). 
Stochastic Variational Gaussian Process (SVGP) optimizes the location and values of a set of pseudo data, usually referred to as inducing points \( \bm{Z} \) and a variational distribution  \( q(\bm{u}) \). 
The parameters of the kernel and the variational parameter of the approximated distribution are optimized using the evidence lower bound (ELBO) as the evidence of the data\cite{titsias2009variational, hensman2013gaussian} . 

The use of SVGP, in the context of this paper, can make the computation significantly faster when fitting a transportation function with many points from the source and target distribution; however, any of the proposed algorithms, formalized in this paper is independent of the approximation nature of the GP. 
 
When dealing with derivatives of transportation maps that are estimated with probabilistic functions, it is necessary to estimate the uncertainty over the predicted derivatives. 
Fortunately, a GP derivative is also a GP
\cite{williams2006gaussian} 
and its existence will depend on the differentiability of the kernel function. The correlation between derivative samples can be expressed as the second partial derivative $k^{11}=\frac{\partial^2}{\partial x_i \partial x_j} k(x_i, x_j)$ while the correlation between derivative samples and function samples is $k^{10}=\frac{\partial}{\partial x_i } k(x_i, x_j)$. 
Thus, the mean and variance prediction of the derivative of the Gaussian Process become   
\begin{gather}
    \begin{aligned}
        \bm{\mu '}&=\bm{K}_{\bm{X}_*,\bm{X}}^{10}(\bm{K}_{\bm{X},\bm{X}}+ \sigma^2_n \bm{I})^{-1} \bm{y} \\
        \bm{\Sigma ' }&=\bm{K}_{\bm{X}_*,\bm{X}_*}^{11} -\bm{K}_{\bm{X}_*,\bm{X}}^{10} (\bm{K}_{\bm{X},\bm{X}}+ \sigma^2_n \bm{I})^{-1} \bm{K}_{\bm{X},\bm{X}_*}^{01}.
    \end{aligned}
    \label{eq:derivative_gp}
    \end{gather}

\vspace{-3em}
\begin{IEEEbiography}
	[{\includegraphics[width=1in,height=1.25in,clip,keepaspectratio]{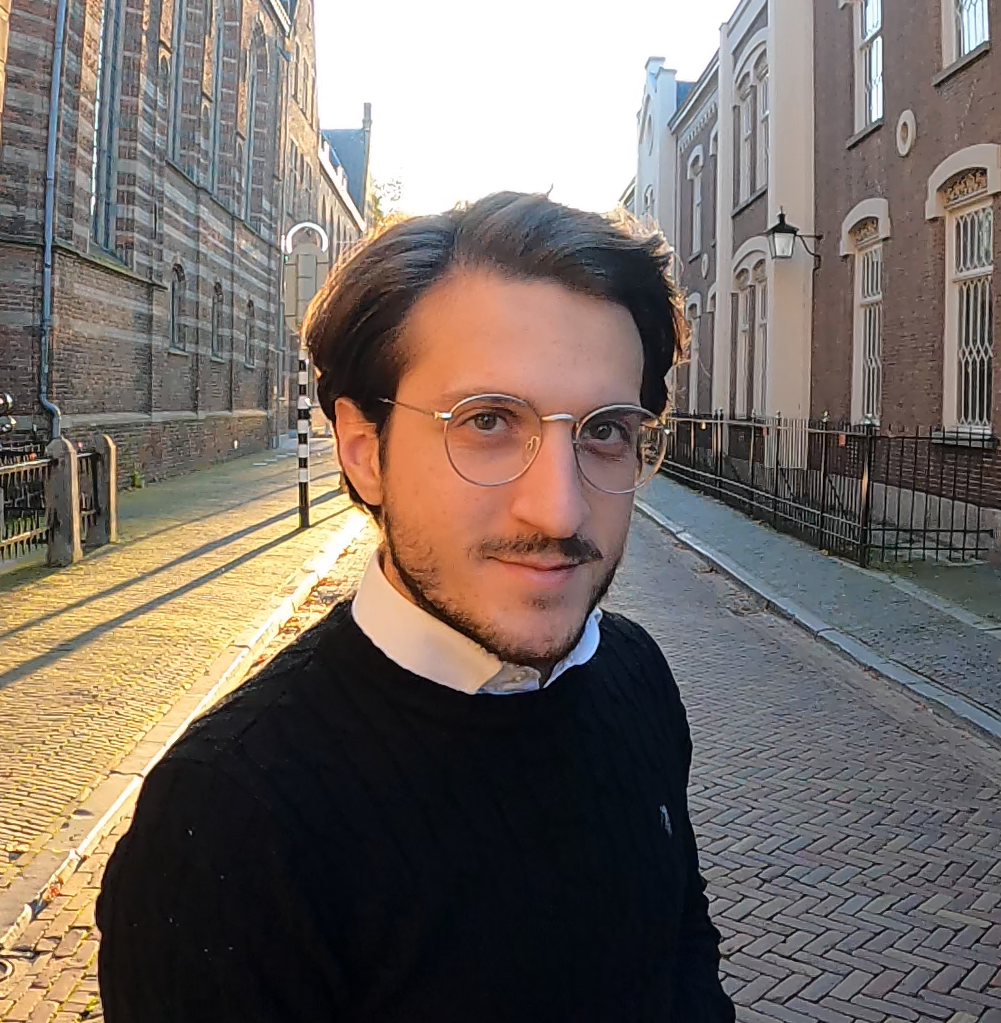}}]{Giovanni Franzese} obtained a Ph.D. in Robotics from the Department of Cognitive Robotics at TU Delft in 2025. He received a BSc degree (2016) in Mechanical Engineering and an MSc degree (2018) in Mechatronics and Robotics at Politecnico di Milano, Italy. He is an ELLIS member for artificial intelligence. His research focuses on Interactive Imitation Learning applied to robot manipulation.  
\end{IEEEbiography}
\vfill
\vspace{-5em} 
\begin{IEEEbiography}
	[{\includegraphics[width=1in,height=1.25in,clip,keepaspectratio]{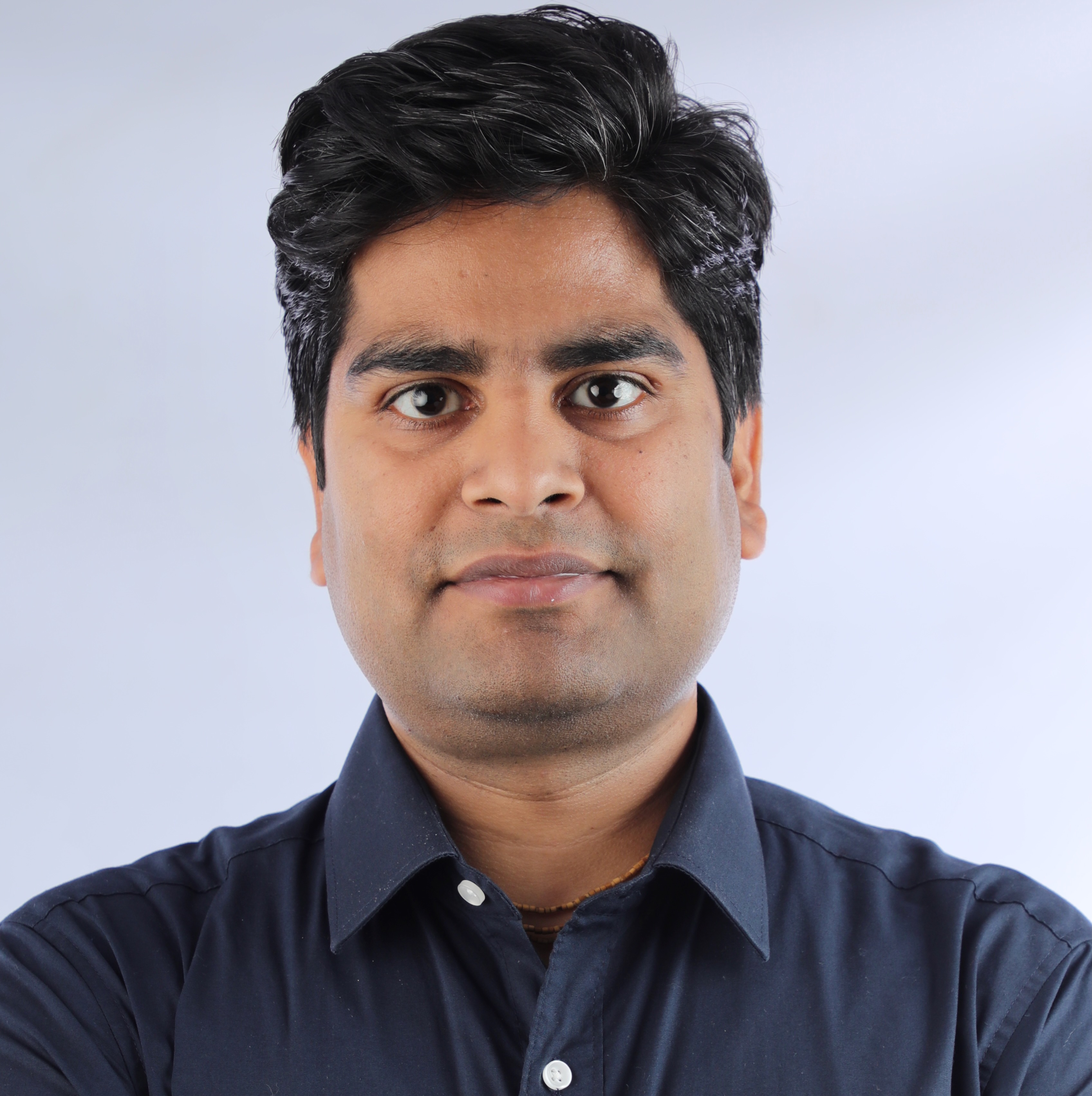}}]{Ravi Prakash}
	is an  Assistant Professor at the department of Cyber Physical Systems (CPS), IISc Bangalore, India. Prior to this, he was a Post-Doctoral researcher in the Learning and Autonomous Control group in the department of Cognitive Robotics, TU Delft. He earned his Ph.D.(2022) in Control and Automation at the Department of Electrical Engineering, Indian Institute of Technology Kanpur. 
 He is a DAAD AInet Postdoc fellow for AI and Robotics.
\end{IEEEbiography} 
\vfill
\vspace{-5em}
\begin{IEEEbiography}
	[{\includegraphics[width=1in,height=1.25in,clip,keepaspectratio]{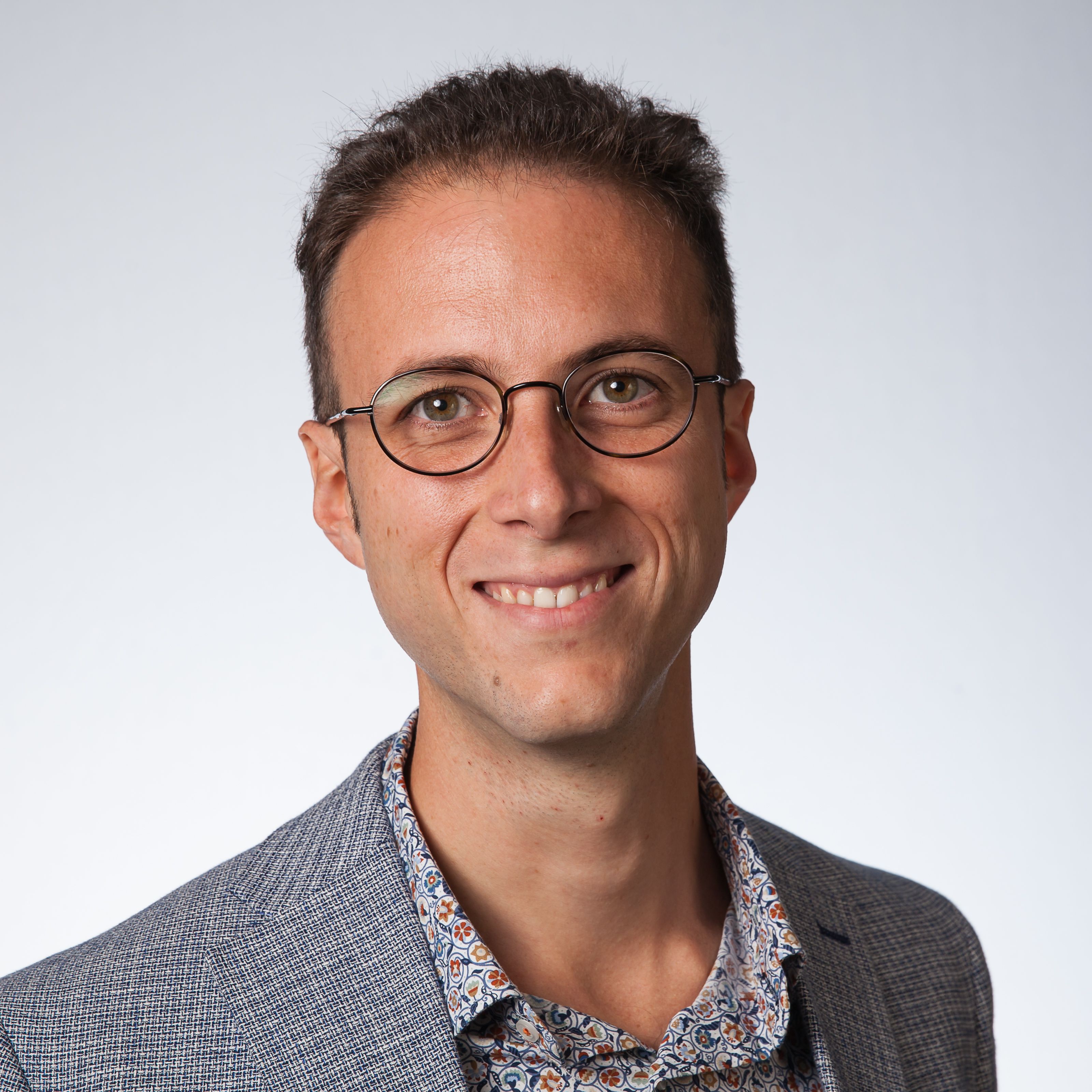}}]{Cosimo Della Santina}
    is currently an Associate Professor with TU Delft, Delft, The Netherlands, and a Research Scientist with the German Aerospace Institute (DLR). He
    received the Ph.D. degree (cum laude) in robotics from the University of Pisa, Italy, in 2019. From 2017 to 2019, he was a visiting Ph.D. student and a Postdoc at the CSAIL at the Massachusetts Institute of Technology. 
    His research interest includes providing motor intelligence to physical systems, focusing on elastic and soft robots. He received the Georges Giralt Ph.D. Award in 2020, the IEEE RAS Early Academic Career Award in 2023, an ERC StG, and an NWO VENI. 
\end{IEEEbiography} 
\vfill
\vspace{-4em} 
\begin{IEEEbiography}[{\includegraphics[width=1in,height=1.25in,clip,keepaspectratio]{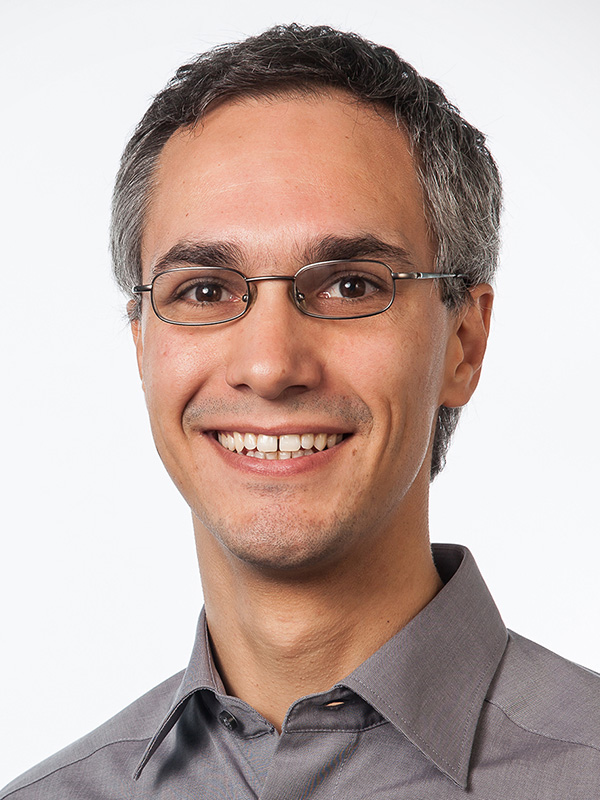}}]{Jens Kober}
	is an associate professor at the TU Delft, Netherlands. He worked as a postdoctoral scholar jointly at the CoR-Lab, Bielefeld University, Germany and at the Honda Research Institute Europe, Germany. He graduated in 2012 with a PhD Degree in Engineering from TU Darmstadt and the MPI for Intelligent Systems. For his research he received the annually awarded Georges Giralt PhD Award for the best PhD thesis in robotics in Europe, the 2018 IEEE RAS Early Academic Career Award, the 2022 RSS Early Career Award, and has received an ERC Starting grant.
\end{IEEEbiography}

\end{document}